%% file: main.tex
\documentclass{article}

    \PassOptionsToPackage{numbers, compress, sort&compress}{natbib}
 \usepackage[preprint]{neurips_2026}

\usepackage[utf8]{inputenc} 
\usepackage[T1]{fontenc}    
\usepackage{url}            
\usepackage{booktabs}       
\usepackage{amsfonts}       
\usepackage{nicefrac}       
\usepackage{microtype}      
\usepackage{xcolor}         

\usepackage{thmtools}
\usepackage{amsmath}
\usepackage{amssymb}
\usepackage{mathtools}
\usepackage{amsthm}
\usepackage{pgfplots}
\usepackage{graphicx}
\usepackage{subcaption}
\usepackage{multirow}
\usepackage{enumitem}
\usepackage{wrapfig}
\usepackage{caption}
\usepackage{siunitx}
\usepackage{standalone}
\usepackage{amsmath}
\usepackage{amssymb}
\usetikzlibrary{shapes, arrows.meta, positioning, shadows, calc, decorations.markings}

\definecolor{cGood}{RGB}{46, 160, 67}
\definecolor{cCheap}{RGB}{31, 119, 180}
\definecolor{cDanger}{RGB}{214, 39, 40}
\definecolor{cStop}{RGB}{255, 127, 14}
\definecolor{cGray}{RGB}{120, 120, 120}

\usepackage{hyperref}       
\hypersetup{
    colorlinks=true,
    linkcolor=magenta,     
    citecolor=blue,     
    urlcolor=blue,       
    pdfborder={0 0 0}    
}

\input{preamble}

\title{Cheap Thrills: Effective Amortized Optimization Using Inexpensive Labels}

\author{%
  Khai Nguyen\\
  Massachusetts Institute of Technology\\
  Cambridge, MA, USA \\
  \texttt{khain@mit.edu} \\
  \And
  Petros Ellinas \\
  Technical University of Denmark \\
  Kongens Lyngby, Denmark \\
  \texttt{petrel@dtu.dk} \\
  \AND
  Anvita Bhagavathula \\
  Massachusetts Institute of Technology\\
  Cambridge, MA, USA \\
  \texttt{abhagava@mit.edu} \\
  \And
  Priya L. Donti \\
  Massachusetts Institute of Technology\\
  Cambridge, MA, USA \\
  \texttt{donti@mit.edu} \\
}

\begin{document}

\maketitle

\begin{abstract}
    To scale optimization and simulation, prior work has explored training machine-learning surrogates that map problem parameters to solutions inexpensively at inference time. Unfortunately, commonly used approaches, including supervised and self-supervised learning with either soft or hard feasibility enforcement, face inherent challenges such as reliance on expensive high-quality labels or difficult optimization landscapes. To address their trade-offs, we propose a novel framework that  collects “cheap” imperfect labels,  performs supervised model pretraining with a merit loss-based termination scheme, and finally refines the model through self-supervised learning to improve final performance. Empirical validation across challenging domains -- including nonconvex constrained optimization, power-grid operation, and stiff dynamical systems -- shows that this three-stage strategy yields faster convergence; improved accuracy, feasibility, and optimality; and up to 59$\times$ reductions in total offline computational cost. We further analyze why and when our framework improves surrogate model training, finding that (i) merit loss is an informative signal and (ii) only small numbers of cheap, inexact labels are needed to place the model in a favorable regime for self-supervised learning.

\end{abstract}

\section{Introduction}\label{sec:intro}

Optimization and simulation are the computational engines behind scientific discovery, engineering design, and operational decision-making. However, classical iterative solvers are often too slow for real-time, high-stakes applications such as power-grid operations, vehicle routing, resource allocation in data centers, and fluid dynamics simulations. To address this, \emph{amortized optimization} (also known as \emph{neural surrogates} or \emph{learning to optimize}) has emerged as a powerful paradigm, which involves training machine learning (ML) models to predict solutions directly from problem parameters, thereby replacing or accelerating expensive iterative solvers with fast forward inference \citep{amos2023tutorial, donti2021dc3, van2025optimization}.

Unfortunately, common approaches to training such surrogate models face a fundamental dilemma. 
While supervised learning (SL) provides stable convergence by regressing onto ground-truth solutions via a simple loss function, generating high-fidelity labels at scale is often prohibitively expensive for complex systems (e.g., large-scale nonconvex optimization and high-order PDEs). 
In particular, label generation requires repeatedly solving the original optimization or simulation problem across many instances, creating a \textit{chicken-and-egg} situation where solving the task is required to avoid solving it.
In contrast, self-supervised learning (SSL) eliminates the need for labeled data by directly minimizing a loss related to the task specification (e.g., the objective and constraints). Although SSL is appealing due to its label-free nature and its direct alignment with the downstream task, the SSL loss landscapes are often highly complex, especially when nonconvex hard constraints or PDE losses are imposed \cite{nguyen2025fsnet, krishnapriyan2021characterizing}. 
Without suitable model initialization, vanilla SSL frequently converges to undesirable local minima, hurting performance. Similar challenges have also been observed in nonconvex optimization \cite{xu2020second}, physics-informed neural networks \cite{daw2022mitigating}, and reinforcement learning \cite{muller2024geometry}.

\begin{figure*}[]
    \centering
    \includegraphics[width=0.9\linewidth]{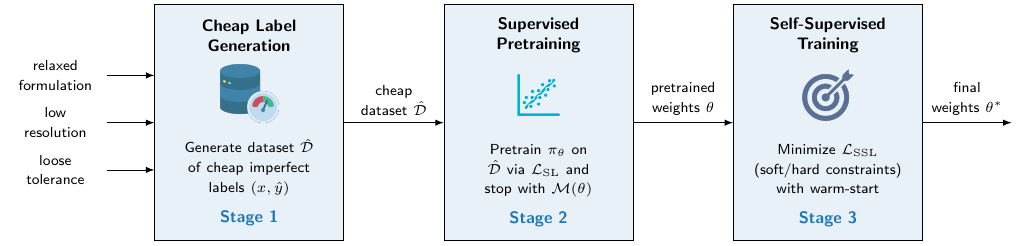}
    \caption{\textbf{Overview of our approach.} We propose a simple but effective three-stage amortized optimization framework:~(1) collecting cheap imperfect labels via approximate solvers, (2) supervised model pretraining using these labels alongside merit loss validation, and (3) self-supervised training.} \vspace{-1em}
    \label{fig:Overview}
\end{figure*}

To address these challenges, we seek to develop an amortized optimization framework that effectively navigates the trade-off between SL and SSL to improve overall solution quality while reducing offline computational costs. 
Our approach is motivated by a simple but theoretically grounded observation: \textit{effective nonconvex optimization requires proper initialization, close to a desirable solution}
\cite{nocedal2006numerical}.
In the context of SSL, this translates to the observation that SSL's success in navigating its complex loss landscape critically depends on how the ML model's parameters are initialized.
To tackle this, we leverage SL to provide a warm start to SSL.
Importantly, however, our SL scheme does not rely on expensive, high-quality labels.
Instead, we use cheap, imperfect labels from approximate optimization or simulation solvers (e.g., iterative solvers with relaxed tolerances or limited iterations, coarse discretizations, or simplified models) together with an evaluation metric called the \textit{merit loss}. 
While low-quality labels are commonly assumed to be of limited utility, our
framework is designed to systematically exploit such imperfect labels as effective initialization signals for SSL. 
We show that careful use of these imperfect labels can indeed be effective in improving SSL performance.

Our contributions are as follows:
\begin{itemize}[topsep=0pt, itemsep=3.5pt, parsep=0pt, partopsep=0pt]
\item We propose an intuitive, effective three-stage framework for amortized optimization: (1) collecting cheap, inexact labels, (2) pretraining our ML model using supervised learning with merit loss-based termination, and (3) training with task self-supervision (Figure~\ref{fig:Overview}).
\item We empirically demonstrate consistent improvements in both performance and training cost 
over SL and SSL baselines under soft and hard constraint enforcement, across challenging nonconvex optimization, AC optimal power flow, and stiff dynamical system domains.
\item Our analysis provides intuition and guides algorithmic design by (i) characterizing how SL with modest numbers of cheap labels suffices to place the model within a favorable regime for SSL, and (ii) showing the informativeness of the merit loss for downstream performance. 
\end{itemize}

\vspace{-0.5em}
\section{Problem formulation}\label{sec:problem}
\vspace{-0.5em}

We study families of continuous constrained optimization problems of the general form
\begin{equation}\label{eq:param_opt_obj}
\min_{y \in \mathbb{R}^n} \; f(y;x)
\quad \text{s.t.} \quad
g(y;x) \le 0,\; h(y;x)=0,
\end{equation}
where $y \in \mathbb{R}^n$ denotes the decision variables and $x \in \mathbb{R}^d$ represents problem parameters. The function
$f : \mathbb{R}^n \times \mathbb{R}^d \rightarrow \mathbb{R}$ defines the objective, while
$g : \mathbb{R}^n \times \mathbb{R}^d \rightarrow \mathbb{R}^{n_{\mathrm{ineq}}}$ and
$h : \mathbb{R}^n \times \mathbb{R}^d \rightarrow \mathbb{R}^{n_{\mathrm{eq}}}$
specify inequality and equality constraints, respectively.

Since the optimal solutions $y^\star$ vary with the parameters $x$, this family of problems induces an implicit mapping from parameters to solutions. Amortized optimization seeks to approximate this mapping using a learned model. Let $\pi_\theta : \mathbb{R}^d \rightarrow \mathbb{R}^n$ denote an ML model parameterized by $\theta \in \mathbb{R}^{n_\theta}$. The learning problem can then be formulated as
\begin{equation}\label{eq:learning_problem}
\min_{\theta \in \mathbb{R}^{n_\theta}} \; \mathbb{E}_{x \sim \mathcal{D}} \big[ f(\pi_\theta(x); x) \big]
\quad
\text{s.t.}
\quad
g(\pi_\theta(x); x) \le 0,\;
h(\pi_\theta(x); x) = 0,
\quad
\forall x \sim \mathcal{D}.
\end{equation}
where $\mathcal{D}$ denotes an unknown distribution over problem parameters and training is performed using a finite dataset sampled from this distribution. The goal is to choose $\theta$ to minimize the expected task objective value while ensuring that the predicted solutions satisfy all constraints and generalize to unseen parameter instances of the same problem family.

\begin{remark}[Exact formulation]
\label{rem:indicator}
The problem in \eqref{eq:param_opt_obj} can be equivalently written as
\begin{equation}\label{eq:opt_equi_merit_loss}
    \min_y \; f(y; x) + I_{\mathcal{F}}(y; x),
\end{equation}
where $I_{\mathcal{F}}$ is an indicator function equal to 0 for $y$ in the feasible set $\mathcal{F}(x)$, 
and $+\infty$ otherwise.
\end{remark}

Accordingly, the learning problem \eqref{eq:learning_problem} can be written as
\begin{equation}\label{eq:learn_equi_merit_loss}
    \min_{\theta} \; \mathcal{L}(\theta)
    = \mathbb{E}_{x \sim \mathcal{D}} \big[ f(\pi_\theta(x);x) + I_{\mathcal{F}}(\pi_\theta(x); x) \big].
\end{equation}

A continuous approximation of \eqref{eq:learn_equi_merit_loss} is given by the quadratic penalty (\textit{merit loss}\footnote{
The term \emph{merit} is inspired by classical constrained optimization, where merit functions combine objective value and constraint violation into a single scalar for line-search and globalization methods~\cite{nocedal2006numerical,bertsekas2014constrained}.
}) formulation
\begin{equation}\label{eq:merit_loss}
    \mathcal{M}(\theta)
    = \mathbb{E}_{x}\!\left[
        f(\pi_\theta(x);x)
        + \rho_{\text{merit}} \, \|c(\pi_\theta(x);x)\|^2
    \right],
\end{equation}
where $c(y;x) := \big[ \max(g(y;x),0),\; h(y;x) \big]$ denotes the stacked constraint residuals and $\rho_{\text{merit}} > 0$ is a penalty parameter. Under standard constraint qualifications and exact penalty conditions, minimizing $\mathcal{M}$ recovers solutions of the original constrained problem \eqref{eq:learn_equi_merit_loss} as $\rho_{\text{merit}} \rightarrow +\infty$ \cite{bertsekas2014constrained,nocedal2006numerical}.

Directly optimizing $\mathcal{L}$ with the indicator $I_{\mathcal{F}}$ (Eq.~\eqref{eq:learn_equi_merit_loss}), or optimizing $\mathcal{M}$ with a very large $\rho_{\text{merit}}$ (Eq.~\eqref{eq:merit_loss}), is highly challenging due to non-smoothness and ill-conditioning. Hence, commonly used ML approaches \textit{replace} these objectives with data-driven losses or smooth, softened approximations (e.g., penalty, barrier, or augmented Lagrangian formulations), whose minimizers converge to those of the original problem at the limit (see Appendix~\ref{app:ssec:indicator}). The accuracy of these approximations depends critically on $\rho$, both during label generation (implicitly through solver tolerances that determine solution quality) and during model training (where it acts as a hyperparameter to enforce feasibility).

\textbf{Supervised learning.}
In data-driven approaches, a parametric model is trained to approximate the solution map by regressing onto labels produced by a numerical solver. Specifically, the ML model parameters $\theta$ are learned by minimizing
\begin{equation}\label{eq:sl_obj}
    \mathcal{L}_{\mathrm{SL}}(\theta)
    = \mathbb{E}_{(x,\hat y)} \big[
        \|\pi_\theta(x)-\hat y\|^2 +  
        \rho_{\text{SL}}\|c(\pi_\theta(x); x)\|^2  + R(\pi_\theta(x))
    \big],
\end{equation}
where $\hat y \approx y^\star(x)$ denotes a solution, a small $\rho_{\text{SL}}$ is used to encourage solution feasibility, and $R(\cdot)$ denotes optional regularization. Despite the strong dependence on data, this approach provides stable, low-variance supervision and yields reliable initializations for downstream optimization, making it a common choice in learning-based optimization and control \cite{amos2023tutorial,piloto2024canosfastscalableneural}.

\textbf{Self-supervised learning.}
Alternatively, ones can avoid using labels by training the amortized model directly on a task-informed loss with standard SGD methods \cite{nguyen2025fsnet}:
\begin{equation}\label{eq:ssl_obj}
        \mathcal{L}_{\mathrm{SSL}}(\theta)
    = \mathbb{E}_{x} \big[
        f(\pi_\theta(x);x) +
        \rho_{\text{SSL}}\|c(\pi_\theta(x); x)\|^2  + R(\pi_\theta(x))
    \big].
\end{equation}
By avoiding labels, this decouples training cost from the complexity of the numerical solver, 
and has been widely explored in differentiable optimization and physics-informed learning
\cite{gould2016differentiating,agrawal2019differentiable,amos2017optnet}.

\begin{wrapfigure}{r}{0.45\textwidth}
    \centering
    \vspace{-1.7em} 
    \includegraphics[width=0.47\linewidth, trim=300 100 300 400, clip]{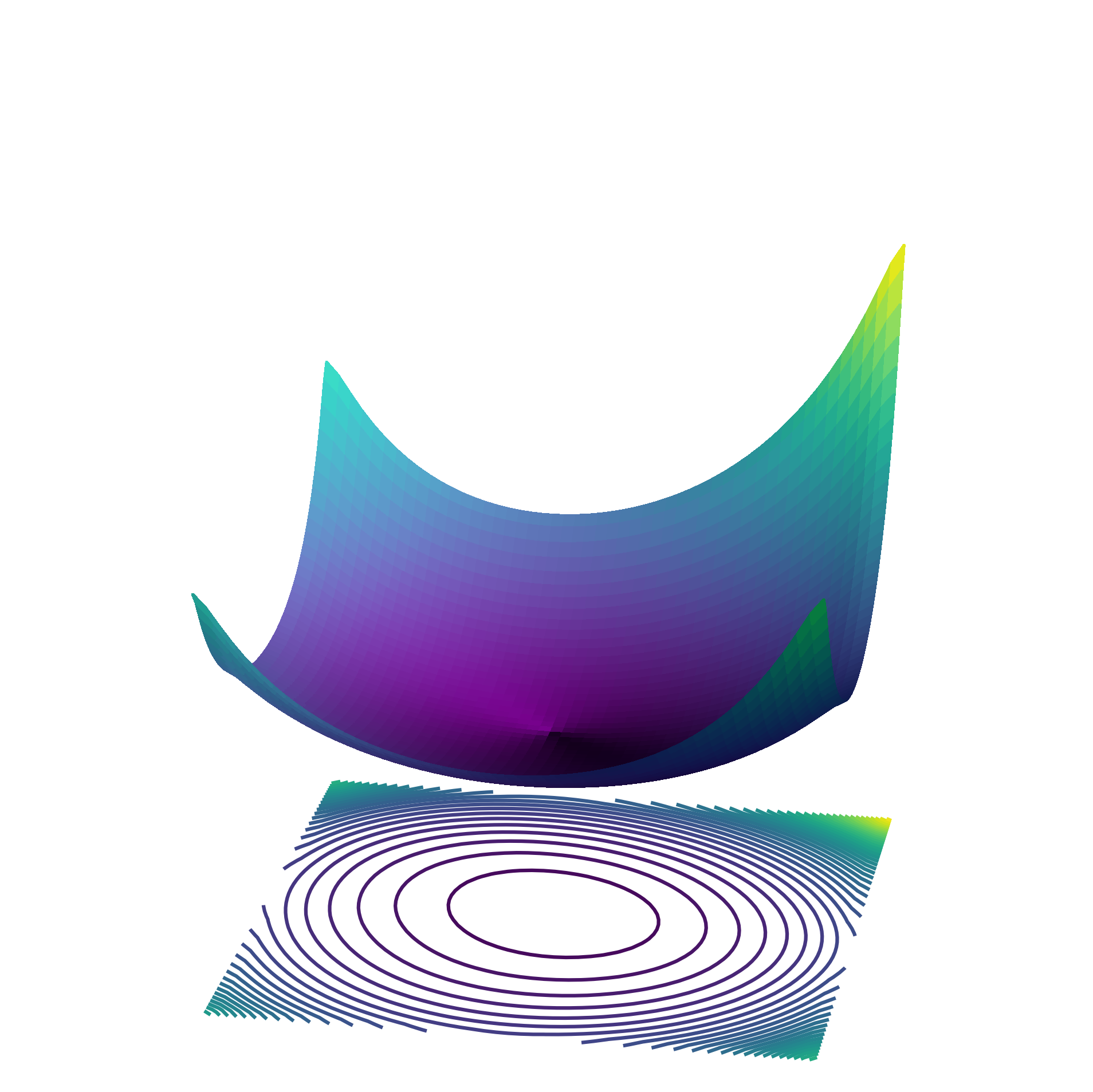}
    \hfill
    \includegraphics[width=0.47\linewidth, trim=300 100 300 400, clip]{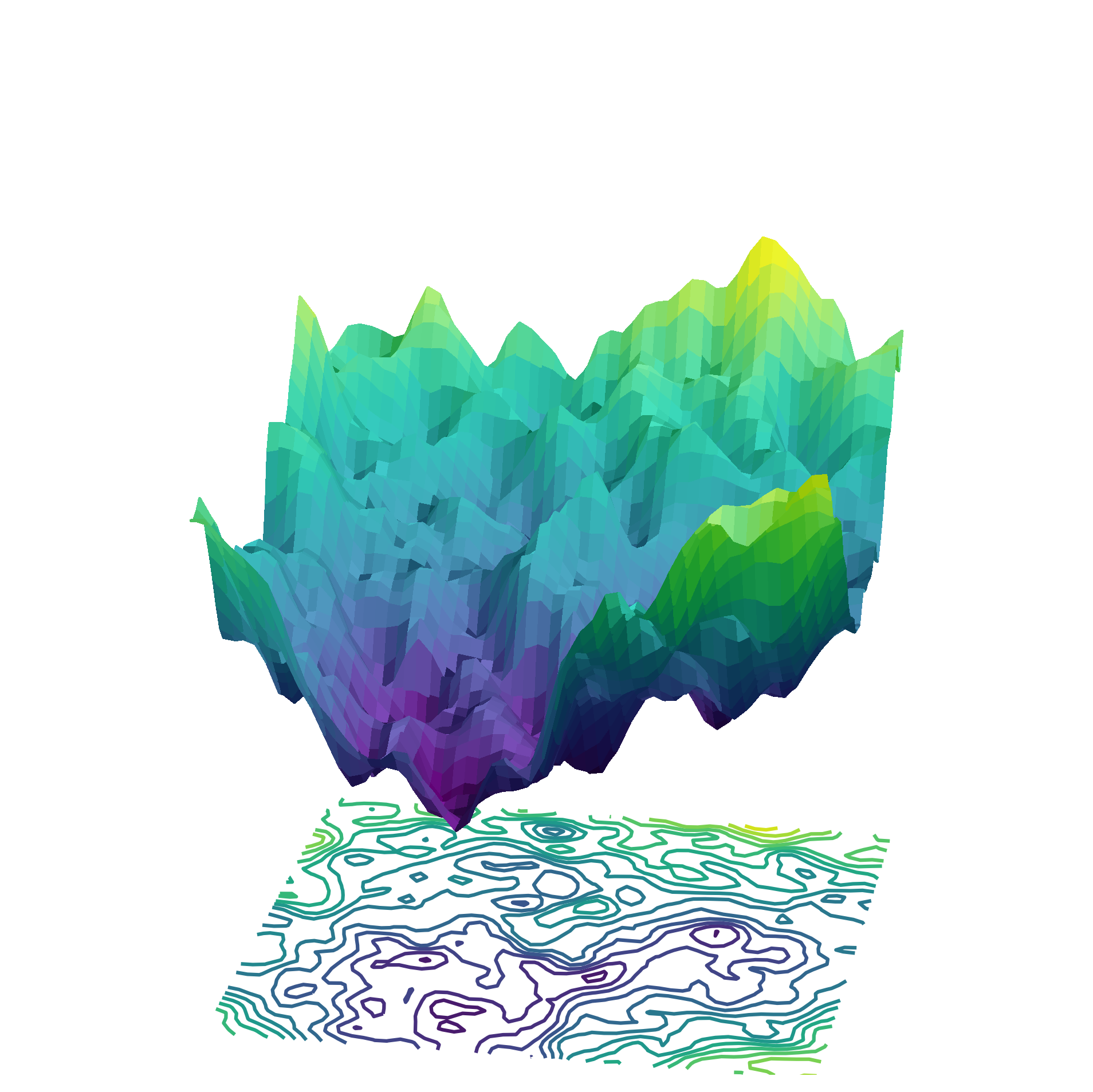}
    \caption{\textbf{Training loss $\mathcal{L}_{\mathrm{SSL}}(\theta) $ (left) and merit loss $\mathcal{M}(\theta)$ (right) landscapes along two weight directions} from a trained FSNet in the nonconvex constrained optimization task.
    The trained weights land in a smooth local minimum of $\mathcal{L}_{\mathrm{SSL}}$. When evaluated with the more task-faithful $\mathcal{M}$, the surroundings are extremely rugged with sharp ridges and multiple local minima. This explains the potential failure of vanilla SSL when trained directly on $\mathcal{M}$. Here, $\rho_{\text{merit}} = 10^4 \rho_{\text{SSL}}$.}
    \label{fig:socp_loss_merit_landscape}
    \vspace{-1em}
\end{wrapfigure}

\textbf{Failure modes and objective mismatch.}
Both SL and SSL rely on surrogate losses \eqref{eq:sl_obj} \& \eqref{eq:ssl_obj} to improve numerical conditioning and facilitate learning of over-parameterized networks ($\rho_{\text{SL}} \approx \rho_{\text{SSL}} \ll \rho_{\text{merit}}$). However, SL critically depends on data, with low-quality labels commonly leading to task failures. On the other hand, stronger feasibility in SL and SSL typically requires increasing \(\rho\). As \(\rho \to +\infty\), the objective becomes increasingly ill-conditioned or non-smooth, making training difficult and often impeding convergence to high-quality solutions. End-to-end hard-constrained approaches such as \cite{donti2021dc3, nguyen2025fsnet} avoid large \(\rho\) by introducing iterative repair layers; however, these components induce complicated training dynamics that can also hinder consistent performance (see Appendix~\ref{app:sec:discussion}). Thus, both soft- and hard-constrained methods face training challenges, albeit for different reasons. While additional regularization or constraint relaxation may ease training, they introduce objective mismatch. As shown in Figures~\ref{fig:socp_loss_merit_landscape}~and~\ref{fig:socp:average_merit_epochs}~(left), training losses may produce more favorable optimization landscapes, but no longer align with the true task-aligned merit loss $\mathcal{M}$ \eqref{eq:learn_equi_merit_loss}, which itself is never directly optimized 
due to challenging landscapes, especially when hard constraints or PDEs are enforced.

\section{Self-supervision from supervised warm-starting via cheap labels}\label{sec:our_method}

The chicken-and-egg problem of SL and the fragility of pure SSL motivate an intuitive, principled strategy that combines the strengths of both.
As shown in Figure~\ref{fig:Overview}, we propose a three-stage pipeline:

\textbf{Stage 1: Cheap label generation.}
We construct a dataset $\hat{\mathcal{D}} = \{(x_i, \hat y_i)\}_{i=1}^N$ of inexpensive labels using a procedure $\hat{\mathcal{S}}$ -- e.g., a numerical solver, sampler, or simulator that leverages approximation techniques such as relaxed solution tolerances, limited iterations, coarse discretizations, or simplified and linearized formulations. The imperfect labels potentially preserve the coarse structure of the solution manifold, while $\hat{\mathcal{S}}$ reduces offline label-generation cost by orders of magnitude compared to using a ``true'' higher-quality solution procedure.

\textbf{Stage 2: Supervised pretraining.}
We pretrain a plain neural network model $\pi_\theta$ 
on $\hat{\mathcal{D}}$ using standard SL with $\mathcal{L}_{\mathrm{SL}}$ defined in \eqref{eq:sl_obj}. The role of this stage is not to achieve high-precision solutions, but rather to provide a favorable initialization for the neural network model. 
Due to the smoothness and stability of the SL objective, this phase converges rapidly. More importantly, instead of monitoring this pretraining phase with  $\mathcal{L}_{\mathrm{SL}}$, we monitor it using the merit loss $\mathcal{M}$, with the \textit{merit-minimum} determining 
when SL training should stop
to avoid overfitting to cheap label bias.

\textbf{Stage 3: Self-supervised training from warm-start.}
Starting from the pretrained parameters, we then apply SSL to minimize $\mathcal{L}_{\mathrm{SSL}}$ in \eqref{eq:ssl_obj}.
In this stage, we can optimize either the plain neural network  model (i.e., with only soft constraint enforcement via the loss), or a version with additional hard constraint enforcement modules appended to it (e.g., \cite{donti2021dc3, nguyen2025fsnet}).
Once the SL-provided ``warm-start initialization'' of the neural network already lies within a task-aligned region, SSL is substantially more stable and converges more reliably to better minimum. This consistently leads to faster convergence and improved final performance compared to vanilla cold-started SSL.

Our pipeline offers a simple yet effective framework for amortized optimization, requiring minimal changes to existing methods while smoothing surrogate model training in challenging optimization and simulation settings.
As we will show in our empirical analysis, despite its simplicity, this method is able to substantially reduce offline cost while preserving the scalability and task fidelity of SSL.

\section{Results}\label{sec:result}

\subsection{Methods and metrics}\label{ssec:setup}

\textbf{Baseline soft constraint methods.}
We train a \mt{Supervised} model to minimize $\mathcal{L}_{\mathrm{SL}}$~\eqref{eq:sl_obj} using  high-quality solver-generated labels;
this represents an upper bound on performance achievable with expensive high-quality data. We further consider SSL methods that softly enforce constraints via 
$\mathcal{L}_{\mathrm{SSL}}$~\eqref{eq:ssl_obj}, including a \mt{Penalty} scheme with fixed $\rho_{\text{SSL}}$ and an \mt{Adaptive Penalty}  scheme with adaptive $\rho_{\text{SSL}}$ \cite{fioretto2021lagrangian};
these approaches optimize a relaxed merit loss 
and do not guarantee exact feasibility.

\textbf{Baseline hard constraint methods.}
We evaluate state-of-the-art SSL methods including \mt{DC3} \cite{donti2021dc3} and \mt{FSNet} \cite{nguyen2025fsnet}, which impose hard constraints on plain neural network predictions through differentiable optimization layers and/or other architectural components.

\textbf{Our three-stage approach.}
Following Section~\ref{sec:our_method}, we pretrain a 
plain neural network using SL on a small number of cheap, inexact labels, and use it to initialize the aforementioned SSL methods.

\textbf{Evaluation metrics.} We report average and worst-case objective values (or optimization gaps relative to a numerical solver IPOPT \cite{biegler2009large}), and $\ell_1$ norms of equality and inequality constraint violations over multiple random seeds.
We use the merit loss $\mathcal{M}$, with large $\rho_{\text{merit}}$ serving as a numerical indicator of feasibility (e.g., $10^5$ for \texttt{float32}). Batched timing evaluations for all learning-based methods are evaluated on a single NVIDIA L40S or H200 GPU, while sequential timing evaluations for all methods (including the numerical solver) are run on Intel Xeon Platinum 8462Y+ CPUs.

\input{figures/table_exp_socp_main}
\begin{figure}[t!]
    \vspace{-0.25em}
    \centering
    \begin{minipage}[t]{0.48\linewidth}
        \centering
        \includegraphics[width=0.9\linewidth, trim=0 -70 0 0, clip]{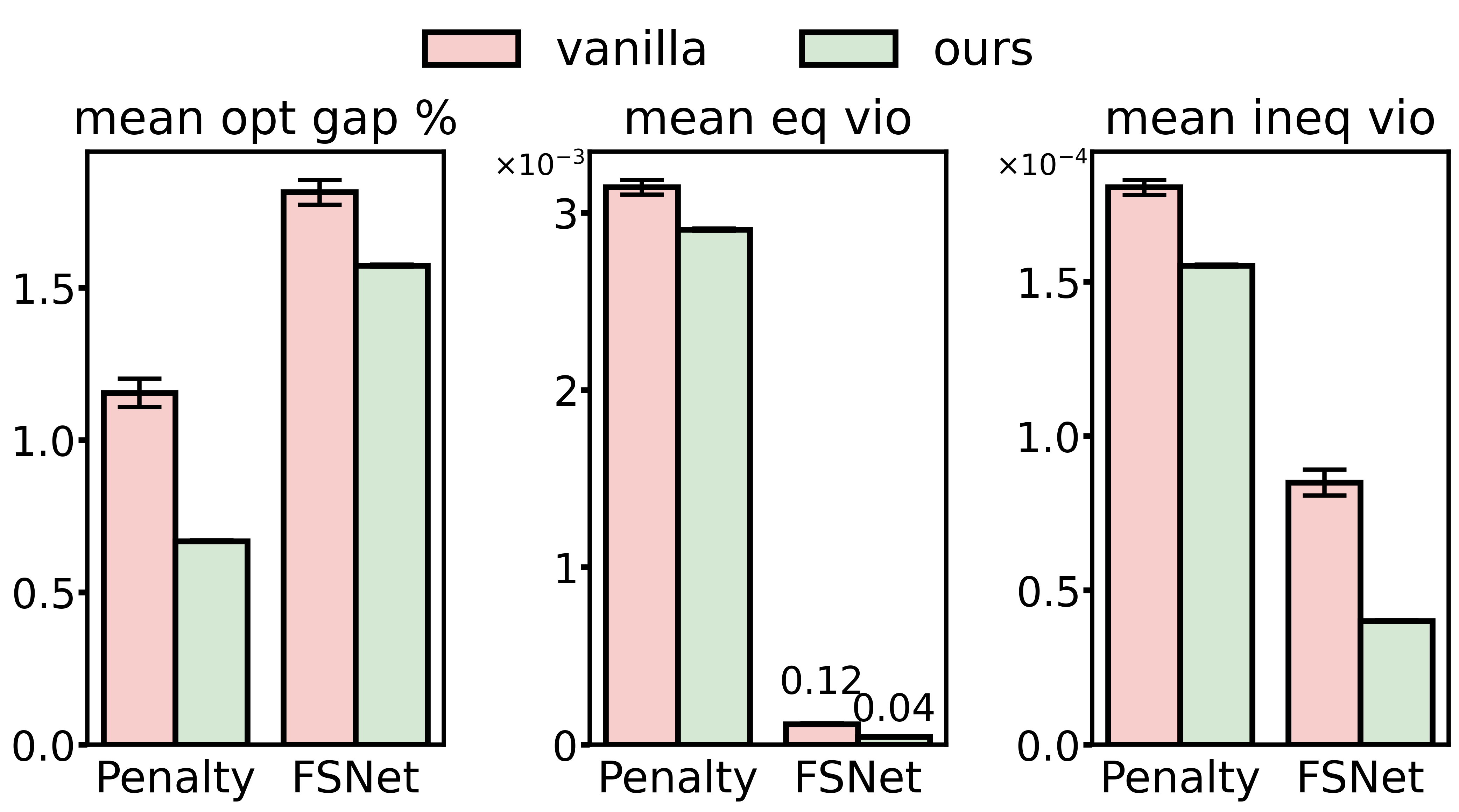}
        \caption{\textbf{
        ACOPF power grid operation.}
        Using cheap DCOPF labels to warm-start SSL 
        reduces average optimality gaps and constraint violations, while remaining competitive in worst-case problems (see Figure~\ref{fig:acopf:main_worst_case}). Gains are especially pronounced for hard-constrained methods.}
        \label{fig:acopf:main}
    \end{minipage}
    \hfill
    \begin{minipage}[t]{0.48\linewidth}
        \centering
        \includegraphics[width=\linewidth, trim=0 10 0 0, clip]{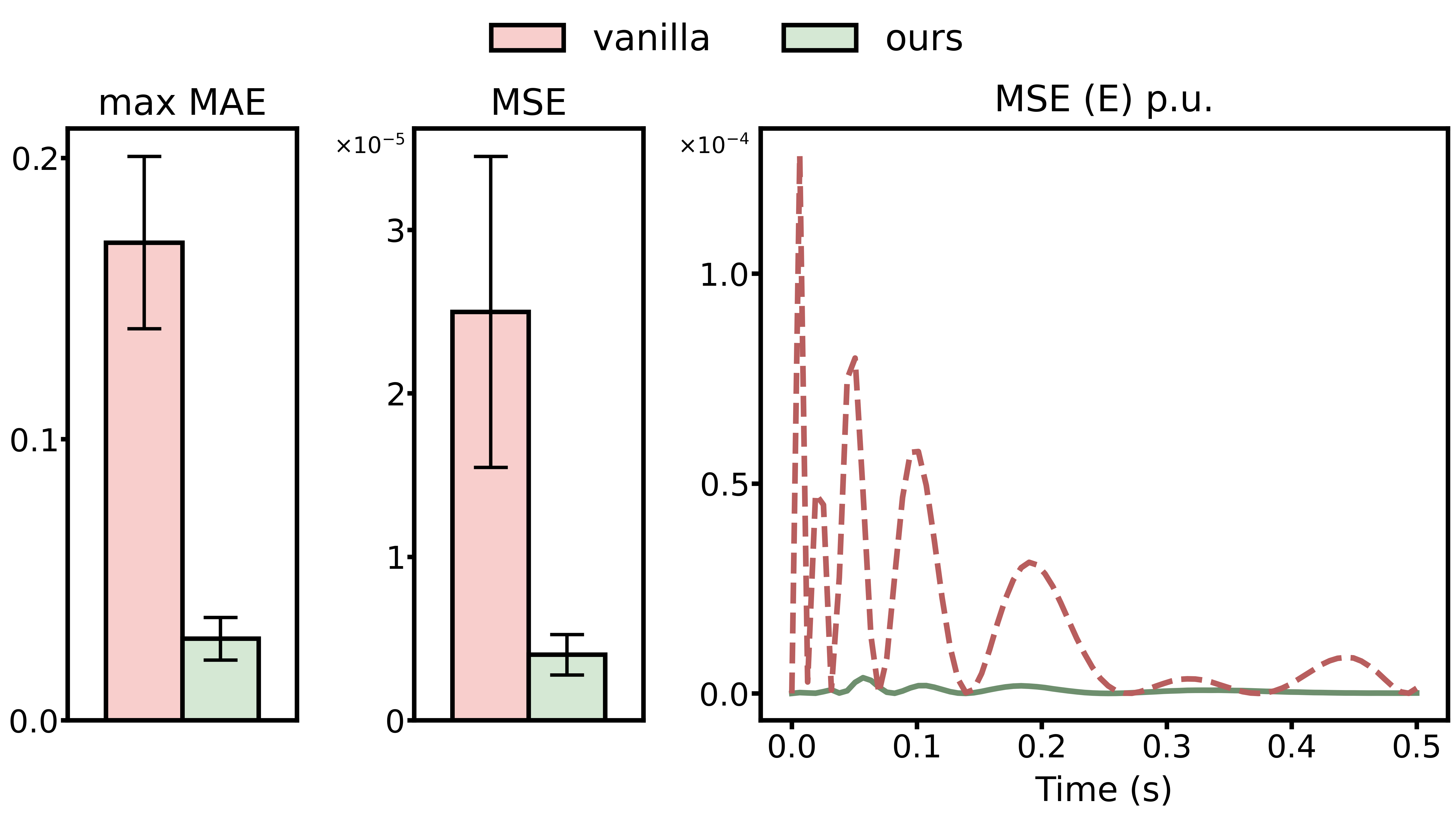}
        \caption{\textbf{Physics-informed learning of a stiff dynamical system.} 
        Left, center: Aggregate solution error 
        relative to ground truth.
        Right: Temporal evolution of error for 
        fastest state ($E$).
        Cheap-label warm-starting lowers aggregate error and suppresses transient spikes in the stiff $E$ state.}
        \label{fig:physics:basin}
    \end{minipage}
    \vspace{-1.em}
\end{figure}

\subsection{Benchmark problems}\label{ssec:benchmark}

\textbf{Synthetic constrained optimization.}\label{ssec:socp}
We first evaluate our method on a challenging parametric optimization problem: a nonsmooth nonconvex second-order cone program  \cite{nguyen2025fsnet}. The problem comprises 100 decision variables subject to 50 equality and 50 inequality constraints. For label collection, we correspond the solution quality for each instance with its maximum CPU solve time or, equivalently, the maximum number of solver iterations (see Appendix~\ref{app:ssec:socp} and Table~\ref{tab:socp:label_quality}). Result comparisons over 2,000 hold-out test instances across methods are reported in Table~\ref{tab:socp:main}, with training dynamics shown in Figure~\ref{fig:socp:average_merit_epochs} and loss and merit loss landscape visualizations in Figures~\ref{fig:socp_loss_merit_landscape}~and~\ref{fig:socp:merit_loss_landscape}. We further conduct empirical analysis on the effect of label quality and label quantity (Figure~\ref{fig:socp:merit_loss_budgets}).

\textbf{Optimal power flow.}\label{ssec:opf}
We consider the practical problem of AC optimal power flow (ACOPF), a nonconvex and NP-hard problem in electric power systems that aims to schedule power generation at minimum cost while satisfying network and device constraints. We adopt the formulation and dataset from \cite{klamkin2025pglearn}. For the IEEE 118-bus ACOPF problem, there are 343 decision variables, 236 equality constraints, and 824 inequality constraints. We acquire 10,000 labels by solving a cheap relaxed formulation, namely DCOPF. Results over 2,000 hold-out test instances are reported in Figure~\ref{fig:acopf:main}; additional details of the formulation and data generation process are provided in Appendix~\ref{app:ssec:opf}.

\textbf{Physics-informed learning.}\label{ssec:physics}
We seek to learn a stiff four-state dynamical system representing electrical grid dynamics,
framed as amortized constrained optimization: a single operator maps an initial condition and timestep to the system state by minimizing weighted physics residuals via automatic differentiation. We compare (i) our approach, which uses cheap linearized data derived from the system Jacobian, against (ii) vanilla physics-only SSL baselines with random initialization. All methods use identical losses, RAR~\cite{wu2023comprehensive} and R3~\cite{daw2022mitigating} adaptive collocation, time-horizon curricula, and second-order polishing. We evaluate convergence reliability and trajectory error against high-accuracy numerical integration (see Figures~\ref{fig:physics:basin}, \ref{fig:physics:basin_r3}, and Appendix~\ref{app:ssec:physics}).

\subsection{Results discussion}\label{ssect:result_dis}
Across all settings, we observe consistent and substantial performance gains from employing our framework. We summarize our 
findings here, with
additional results 
in Appendices~\ref{app:sec:results}~and~\ref{app:sec:discussion}.

\textbf{Our training framework consistently yields better final solutions, outperforming vanilla baselines across all evaluated problem settings}. On synthetic constrained optimization (Table~\ref{tab:socp:main}), our variants achieve lower objective values than their corresponding vanilla
baselines while maintaining or improving feasibility. For example, \mt{Our Penalty} improves the mean objective from \mt{Penalty}'s \(-0.06\) to \(-3.29\) and reduces inequality violations, while \mt{Our FSNet} improves the mean objective from \mt{FSNet}'s \(-0.73\) to \(-3.28\) with lower equality and inequality violations. On ACOPF (Figure~\ref{fig:acopf:main}), warm-starting SSL with cheap DCOPF labels improves both soft- and hard-constrained methods, 
substantially reducing mean optimality gap and constraint violations for hard-constrained methods.
In physics-informed learning (Figure~\ref{fig:physics:basin}), our method reduces aggregate MAE/MSE and stabilizes temporal trajectories: vanilla SSL exhibits large transient error spikes, whereas our warm-started model keeps the fastest state variable \(E\) (the most unstable one) close to the ground-truth trajectory.

\input{figures/table_exp_offline_time}

\textbf{Our strategy reduces total offline cost despite additional stages.}
In Table~\ref{tab:offline_time}, we compare the offline computational cost required by our strategy and its vanilla counterpart to achieve equivalent levels of performance.
Our strategy 
consistently 
reduces the offline cost of the hard-constrained FSNet method, with a $1.7\times$ reduction in the ACOPF task (where SSL training cost dominates).
For soft-constrained methods, while our approach incurs additional cost relative to vanilla \mt{Penalty}, it reduces total offline time compared to the SL baseline by 
$59\times$ 
in the constrained optimization task (where high-quality label generation cost dominates).
These savings arise because imperfect labels substantially reduce both label-generation overhead and the number of SSL training steps. Moreover, when imperfect measurements or historical data are already available, we can skip data generation and still effectively reduce training cost. Finally, our method inherits the advantages of amortized optimization methods, achieving orders-of-magnitude speedups in solution time at inference 
compared to the iterative solver (Table~\ref{tab:socp_soln_time}). Overall, our framework is not only effective but also computationally efficient.

\subsection{Behavior analysis}\label{ssec:behavior}

To better understand our framework, we conduct comprehensive empirical studies to analyze
several 
 questions: (1) How cheap can labels be while  remaining useful for improving SSL? (2) Are large numbers of labels needed to compensate for low data quality? (3) Can random labels also work within our framework? (4) How do cheap-label warm starts affect SSL training dynamics? (5) Is merit loss an informative signal for monitoring and evaluating amortized models on downstream tasks?

\begin{wrapfigure}{r}{0.52\textwidth}
    \centering
    \includegraphics[width=0.49\linewidth]{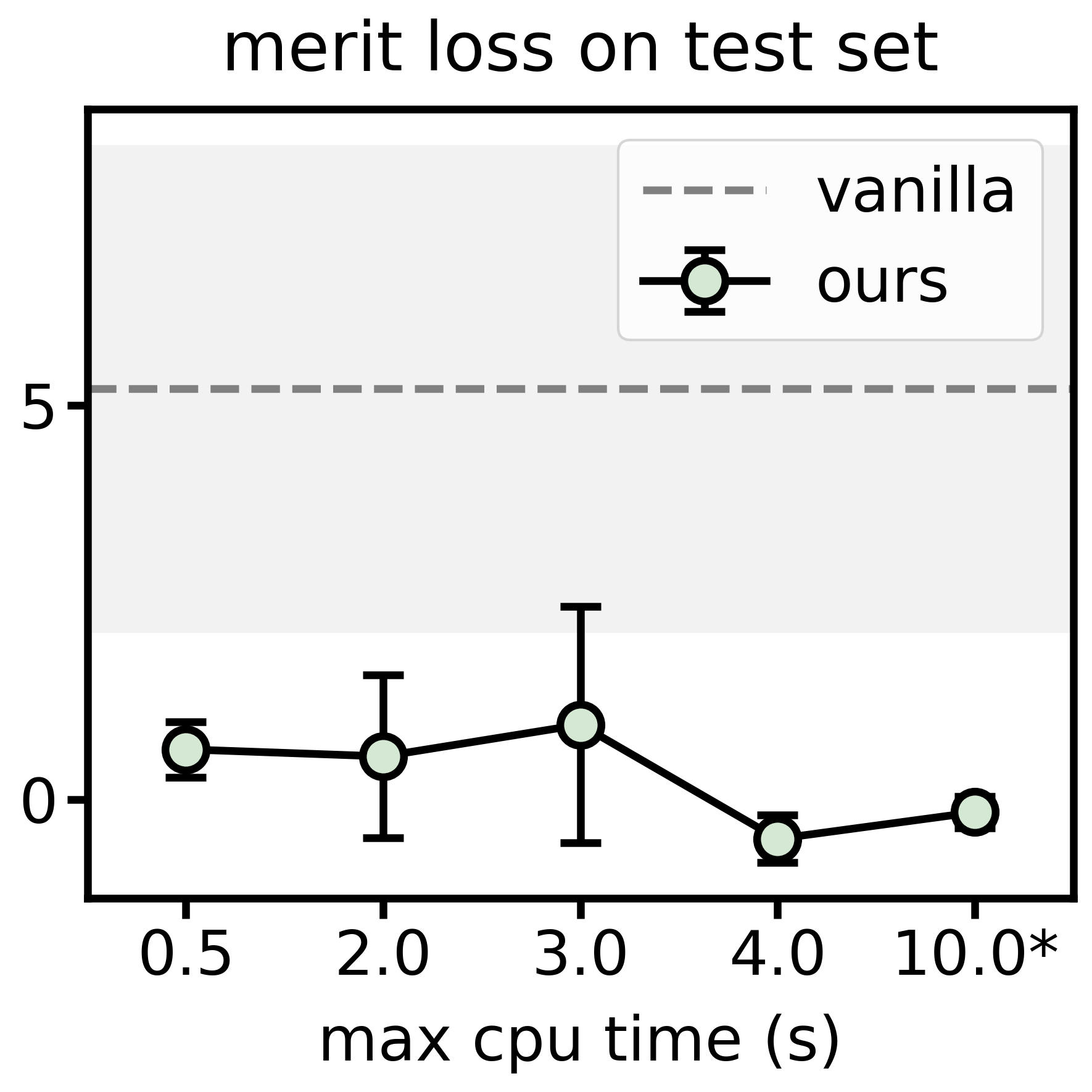}
    \includegraphics[width=0.49\linewidth]{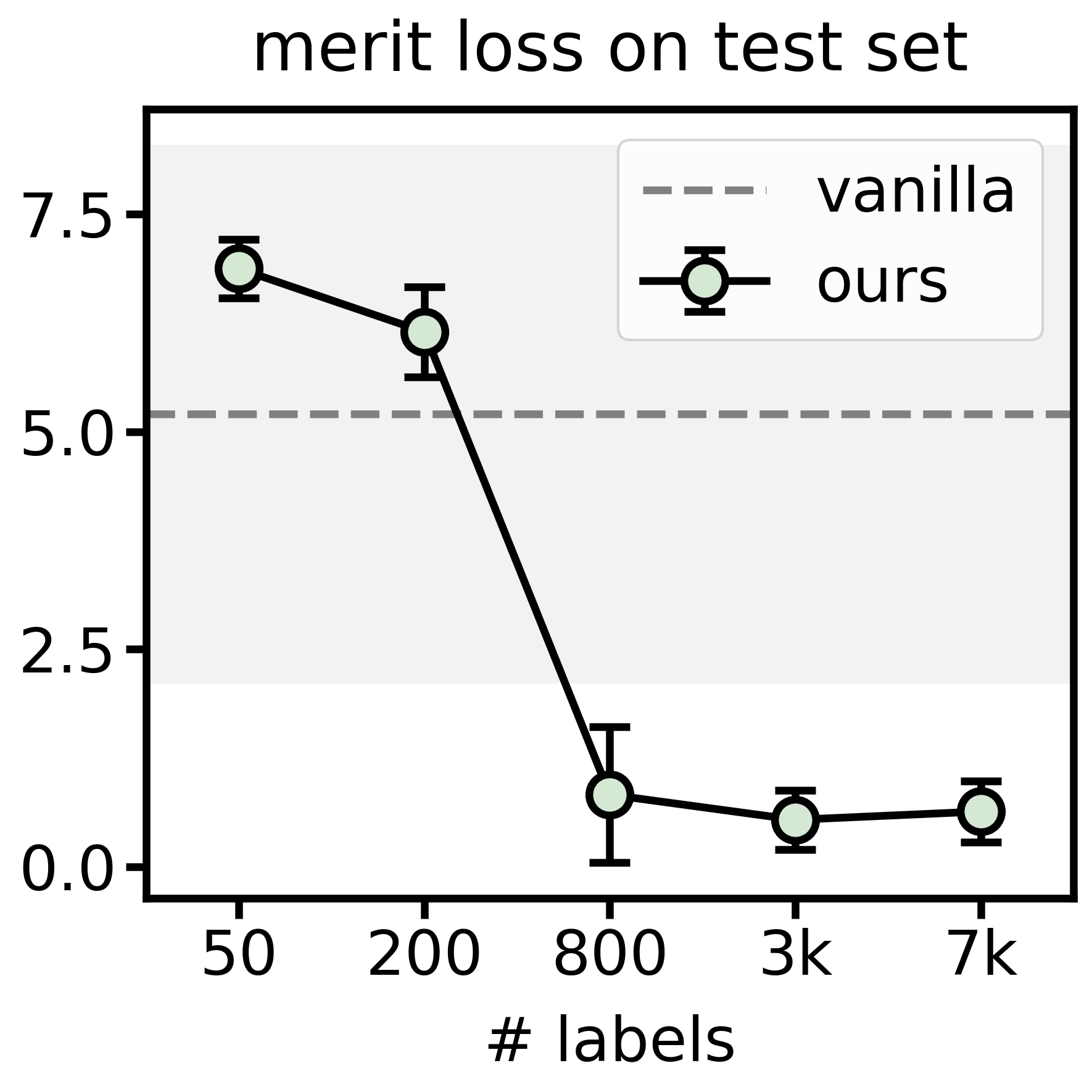}
\caption{\textbf{Performance of our method across imperfect label qualities and quantities.} 
Left: Across different levels of label refinement (with 10~s max CPU time per instance allowing the oracle solver to converge on all instances), performance remains statistically indistinguishable, indicating that higher-accuracy labels provide only marginal additional benefit within our framework.
Right: Performance improves rapidly with a small number of labels and then plateaus, indicating that coarse manifold structure is captured early and is sufficient for subsequent SSL.
}
    \vspace{-1em}
    \label{fig:socp:merit_loss_budgets}
\end{wrapfigure}

\textbf{(1) Extremely cheap labels are sufficient to improve self-supervised training.}
In Figure~\ref{fig:socp:merit_loss_budgets}~(left), in the synthetic optimization benchmark, we vary the maximum CPU time allocated to the solver $\hat{\mathcal{S}}$ for generating each label, from 0.5\,s to 10.0\,s per instance. Larger values result in more refined labels, while smaller values return lower-quality labels. As shown in Table~\ref{tab:socp:label_quality}, these labels differ by orders of magnitude in objective value, constraint violation, success rate, and solver merit. Nevertheless, the final SSL performance remains statistically similar across all budgets. This suggests that once the labels are sufficiently informative to guide the model into a useful regime for SSL convergence, further improving label quality provides only marginal benefit. Similar trends appear across domains among \textit{different protocols} of collecting cheap labels. In power grid operation, DCOPF labels, despite lower equality feasibility, are generated $25\times$ faster than ACOPF labels thanks to the simplified formulation (Table~\ref{tab:acopf:label_quality}), while still improving downstream feasibility and optimality after SSL (Figure~\ref{fig:acopf:main}). In physics-informed learning, linearized dynamics provides $100\times$ cheaper data (Figure~\ref{fig:physics:combined_analysis}, right), which help reduce errors and stabilize simulation trajectories (Figure~\ref{fig:physics:basin}).

\textbf{(2) Only small numbers of labels are needed to provide structured, effective warm starts.}
As shown in Figure~\ref{fig:socp:merit_loss_budgets} (right) for the synthetic optimization benchmark, performance improves rapidly with the number of labels and then saturates, indicating that the coarse structure of the solution manifold is captured early. As additional labels are added, the data set increasingly includes previously unseen hard instances, leading to a lower merit loss and potentially more robust performance. Our framework achieves significant solution improvements with as few as 800 inaccurate labels, while SL baselines with $10\times$ more labels still fall short (Table~\ref{tab:socp:main}). Having \emph{some} labels is indeed helpful; Table~\ref{tab:socp:fsnet_from_penalty} shows that warm-starting the hard-constraint SSL method without labels (i.e., using a penalty method for the warm start) results in worse average optimality and feasibility.

\input{figures/table_socp_random_labels}

\textbf{(3) Cheap, but not arbitrary: Random labels contain no useful structure for SSL initialization.}
As shown in Table~\ref{tab:socp:ssl_random}, warm-starting SSL with random labels performs worse than vanilla SSL without warm-starting, and substantially worse than our method using cheap labels that preserve structure from the approximate solution procedure. This confirms that random labels fail to provide meaningful geometric guidance toward a favorable regime, whereas a few structured cheap labels from approximate procedures provide useful initialization for subsequent SSL.

\begin{figure}[t!]
    \centering

    \begin{minipage}[t]{0.48\linewidth}
        \centering
            \includegraphics[width=0.49\linewidth]{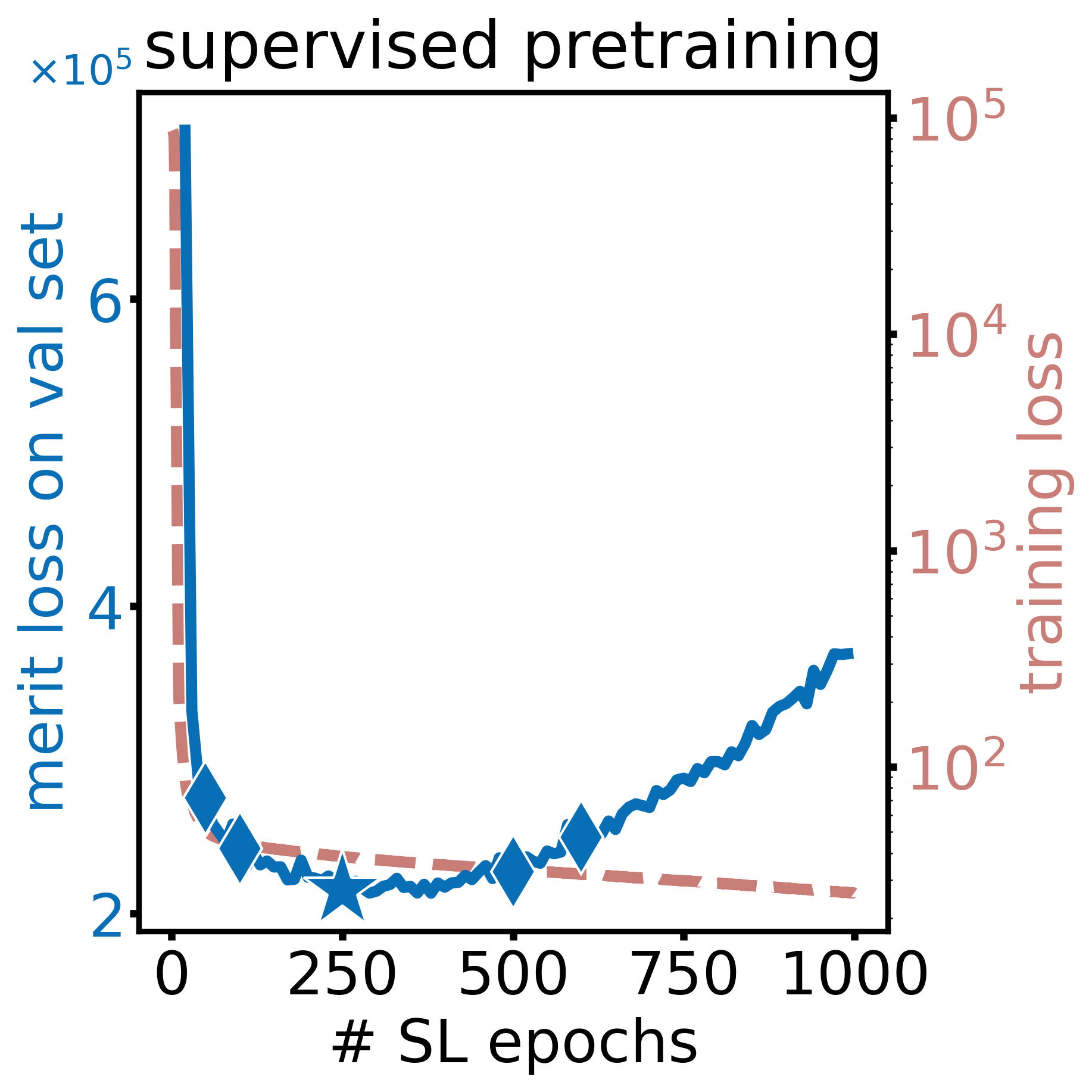}
            \includegraphics[width=0.47\linewidth]{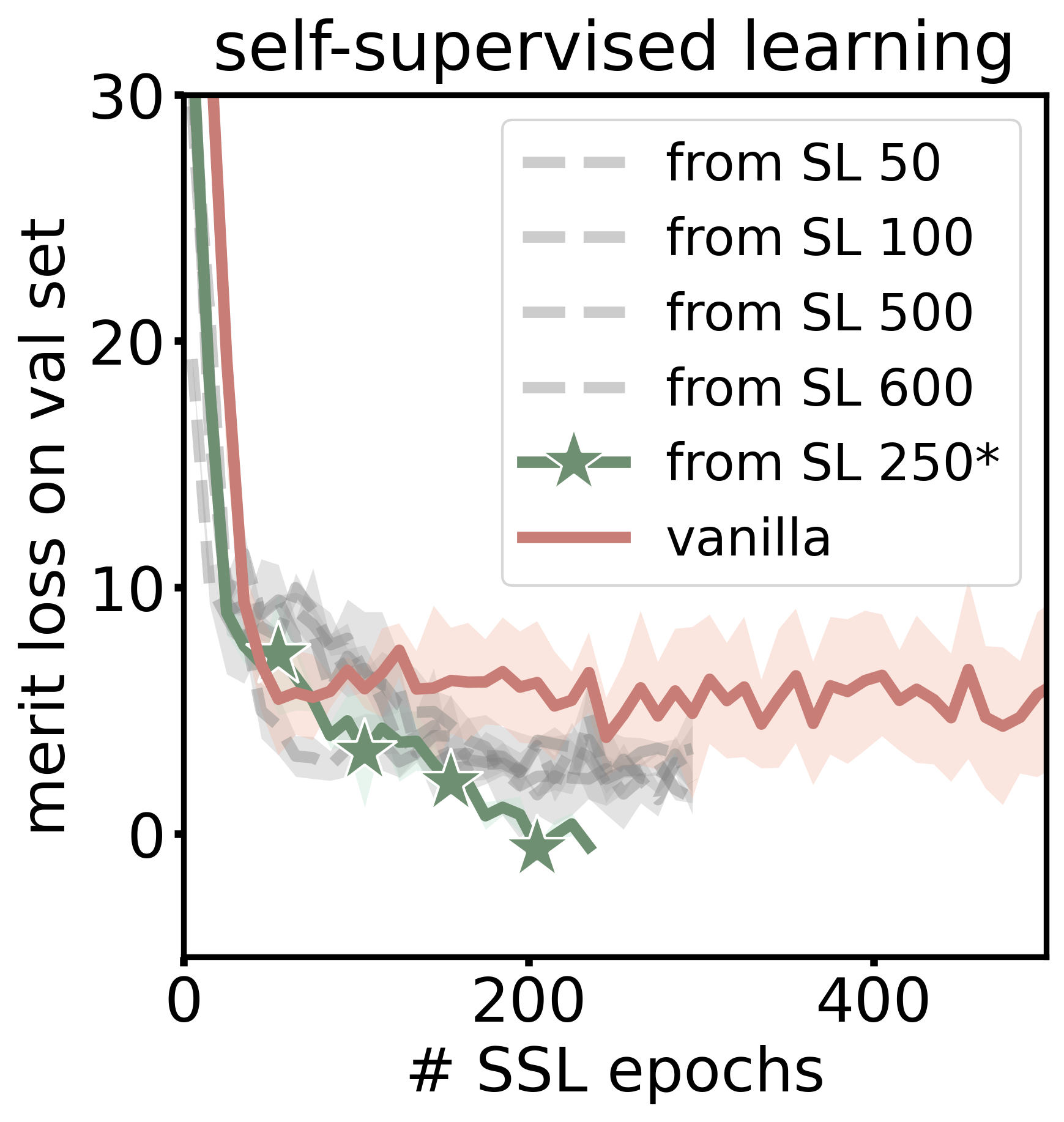}
            \caption{\textbf{Training dynamics over epochs for supervised pretraining, vanilla SSL, and our SSL (from different warm-starts).}
            The SL merit loss (\textcolor{blue}{blue}) follows a U-shaped trajectory whose minimum determines the ideal start of our SSL (at epoch 250), whereas the SL training loss monotonically decreases. Compared to vanilla SSL, which starts from random initialization and then plateaus at higher merit loss, our approach yields better solutions quickly and consistently. }
            \label{fig:socp:average_merit_epochs}
    \end{minipage}
    \hfill
    \begin{minipage}[t]{0.48\linewidth}
        \centering
            \includegraphics[width=0.48\linewidth]{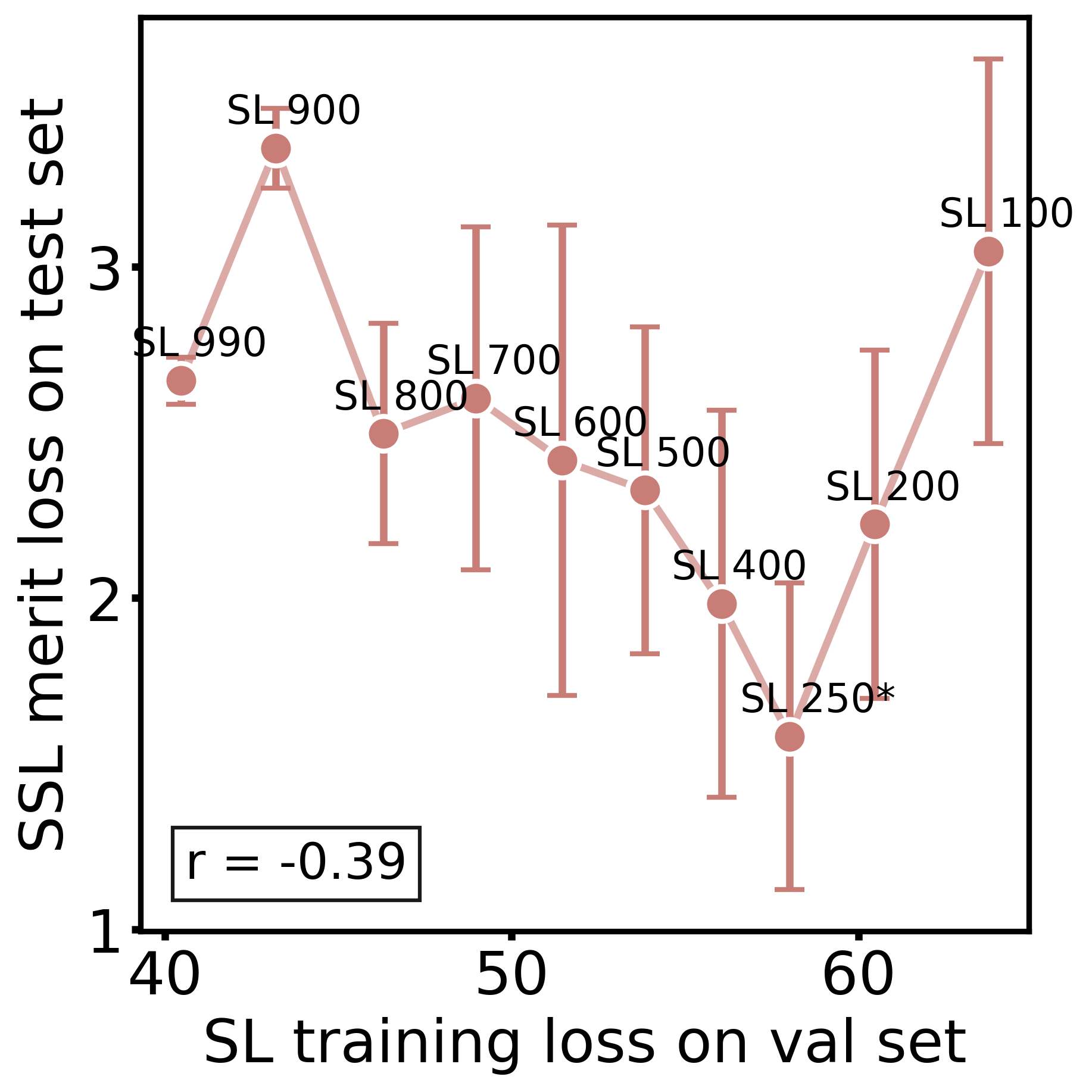}
            \includegraphics[width=0.48\linewidth]{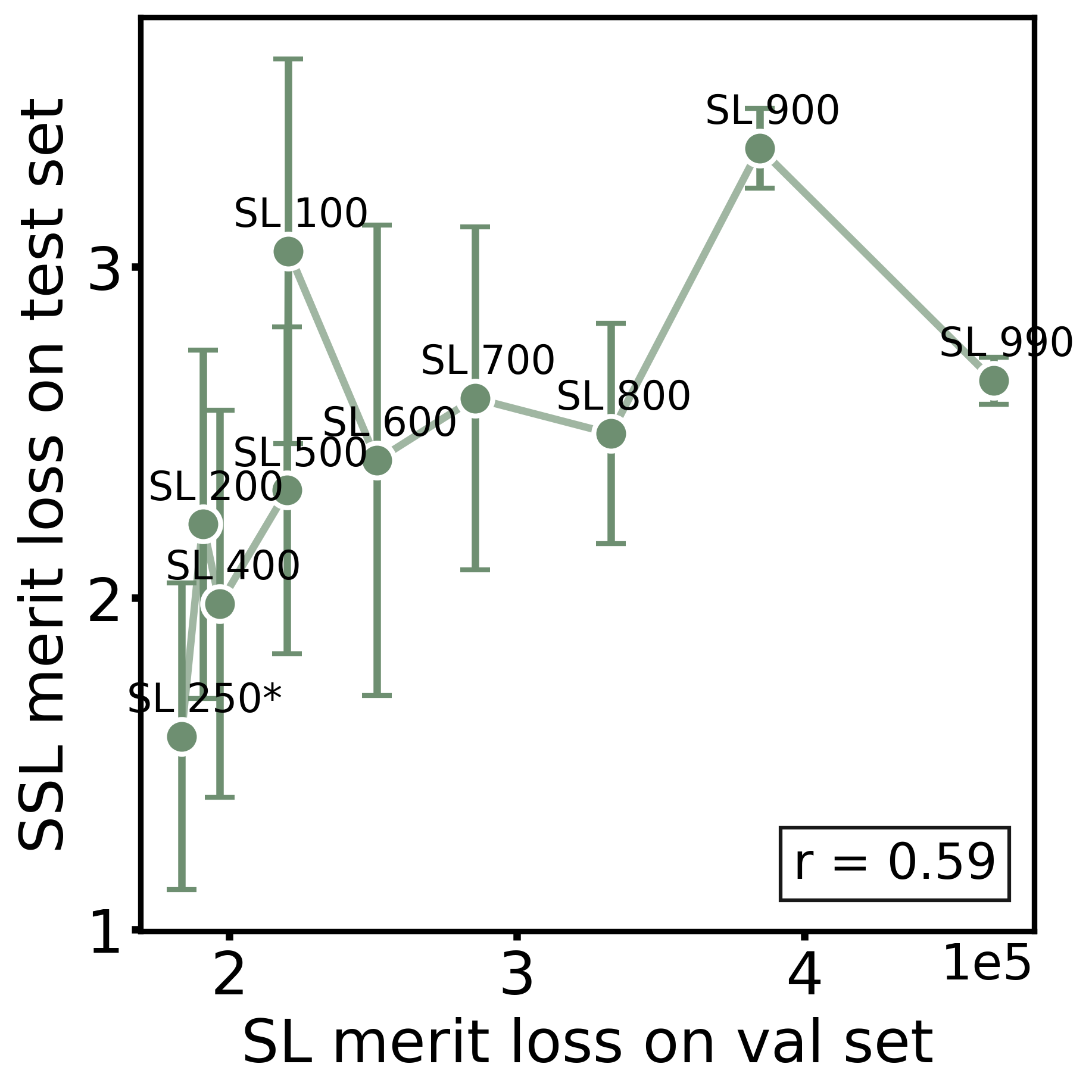}
\caption{\textbf{
Final task performance (SSL merit loss) vs.~SL training and merit losses during pretraining.} 
Left: Lower SL training loss does not consistently improve task SSL performance, suggesting that cheap label overfitting  may degrade SSL initialization quality. Right: Lower SL merit loss correlates with improved downstream task SSL performance, supporting merit loss-based selection. $\mathrm{SL}\ 250^\ast$ corresponds to the SL epoch with the lowest merit loss.
}
            \label{fig:socp:merit_loss_correlation}
    \end{minipage}

    \vspace{-0.5em}
\end{figure}

\begin{figure}
    \centering
    \includegraphics[width=0.24\linewidth, trim=0 -50 0 0, clip]{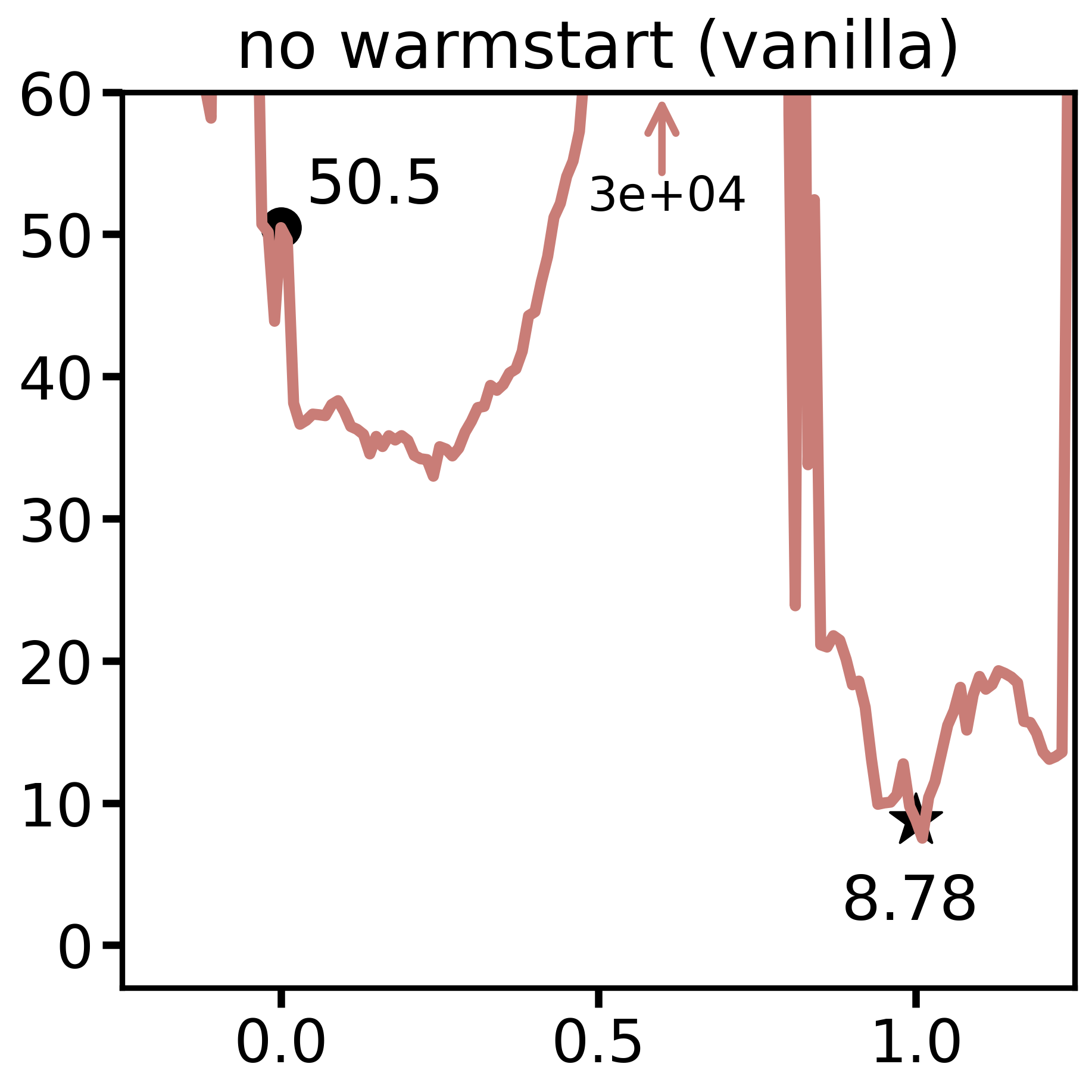}
    \includegraphics[width=0.24\linewidth, trim=0 -50 0 0, clip]{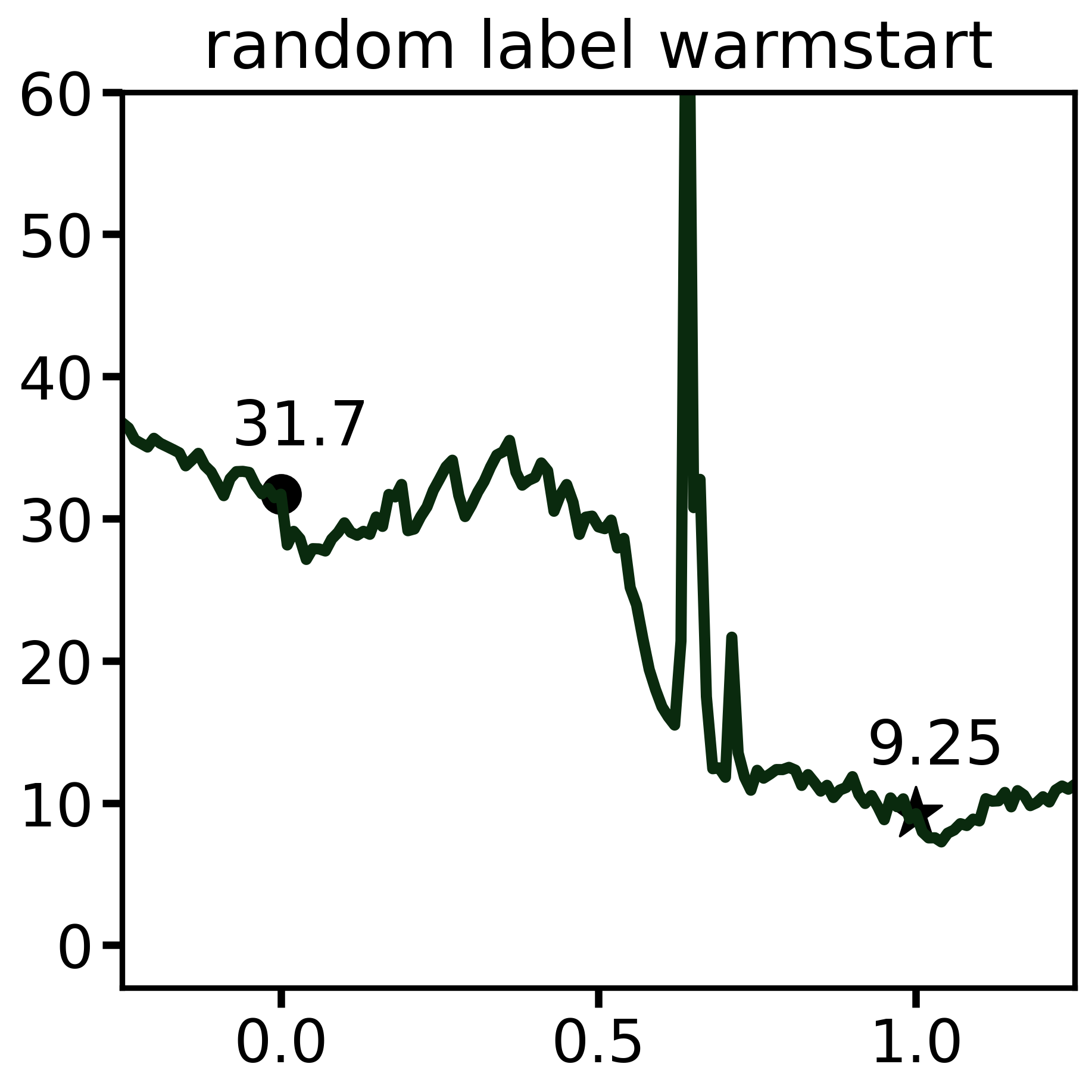}
    \includegraphics[width=0.24\linewidth, trim=0 -50 0 0, clip]{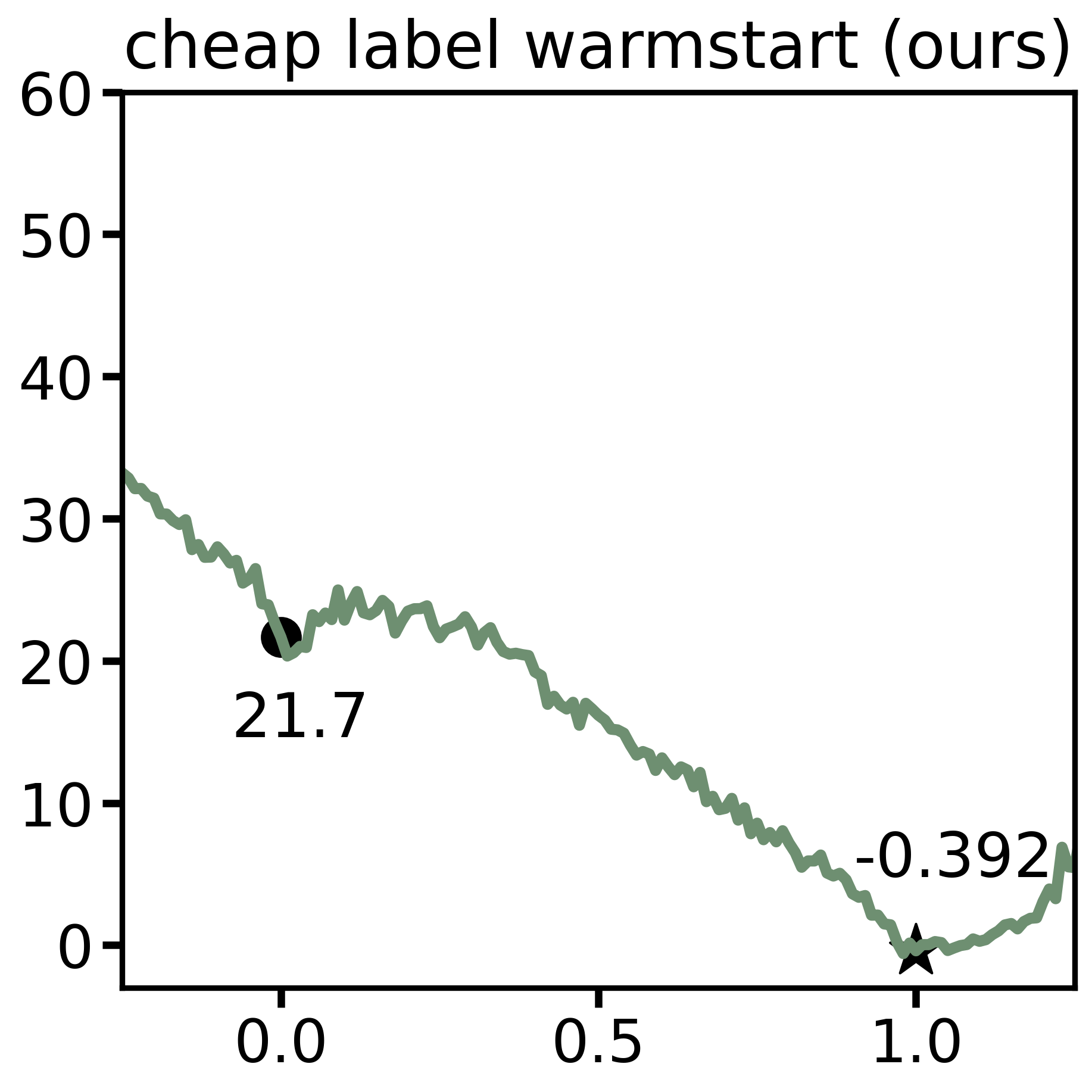}
    \includegraphics[width=0.24\linewidth, trim=0 -50 0 0, clip]{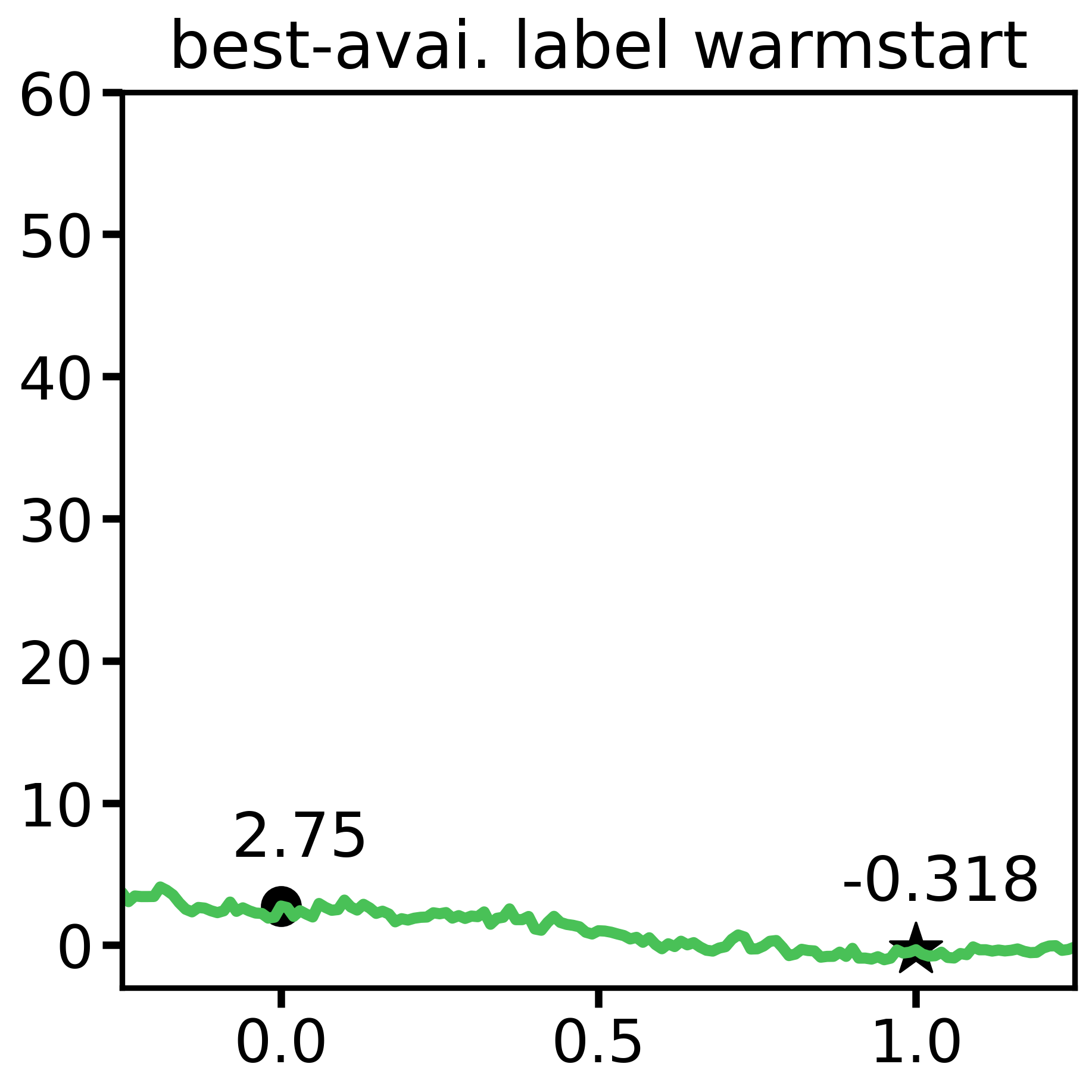} \\
    \caption{\textbf{Merit loss landscapes along linear interpolation between initial ($\alpha=0$) and final ($\alpha=1$) trained weights for different SSL approaches.} Vanilla SSL (no warm-start, random initialization) and SSL warm-started from random labels encounters a rugged landscape with sharp barriers, while SSL warm-started from cheap labels or from best-available high-quality labels shows a smoother, near-monotonic descent toward far better minima. Importantly, warm-starting with more expensive, accurate labels does not show better merit loss than our cheap label strategy.}
    \label{fig:socp:merit_loss_landscape}
    \vspace{-1.5em}
\end{figure}

\textbf{(4) Cheap-label warm starts place the model in a favorable SSL landscape, leading to better training dynamics.}
Figure~\ref{fig:socp:average_merit_epochs}~(right) provides insight into per-epoch training dynamics: while vanilla SSL progresses slowly or becomes trapped at suboptimal plateaus, regardless of the number of SSL epochs, our warm-started SSL quickly converges to high-quality solutions with lower variance.
Figure~\ref{fig:socp:merit_loss_landscape} visualizes merit-loss landscapes along linear interpolations between the start and end points of SSL training. SSL models trained from scratch and random-label warm starts encounter sharp barriers and converge to high-merit-loss solutions, whereas models initialized with informative labels are exposed to a more favorable straight path toward lower-merit (higher-quality) solutions.
We again observe only marginal final performance differences between cheap labels and best-available solver labels; in fact, best labels can bias the model toward a flat local minimum, whereas cheap labels can leave some slack for SSL to explore, with greater momentum, a potentially lower-merit regime.

\textbf{(5) Merit loss is an informative signal for monitoring and evaluating amortized models.}
\(\mathcal{M}\) provides a scalar metric that more faithfully reflects task performance than the training losses \(\mathcal{L}_{\mathrm{SL}}\) and \(\mathcal{L}_{\mathrm{SSL}}\); it also offers a concise alternative to reporting optimality and feasibility separately, as these can evolve in complex and non-monotonic ways. During SL pretraining, \(\mathcal{L}_{\mathrm{SL}}\) decreases monotonically as the model fits the biased labels \(\hat y\), whereas \(\mathcal{M}\) exhibits a characteristic U-shaped trajectory (Figure~\ref{fig:socp:average_merit_epochs}, left), suggesting that the SL path may enter and later exit a task-aligned regime. More interestingly, lower SL merit correlates with improved downstream SSL performance, while lower \(\mathcal{L}_{\mathrm{SL}}\) does not consistently do so (Figure~\ref{fig:socp:merit_loss_correlation}). Our merit-based termination scheme therefore selects the point along such trajectory that is most favorable for subsequent SSL, leading to faster and more stable convergence and improved final solutions after SSL (Figure~\ref{fig:socp:average_merit_epochs}, right). This (i) helps prevent SL from overfitting to cheap, inaccurate labels and (ii) allows better generalization to the task performance.

\subsection{Caveats and Limitations} \label{ssec:limit}
First, our framework does not require the SL merit loss curve to exhibit a U-shape; the transition to SSL is determined by the SL checkpoint with the lowest validation merit loss. Depending on label quality, initialization, and problem structure, the best transition point may occur at the beginning, middle, or end of SL training. We note that label quality is in some cases controllable (e.g., via the approximation procedure), but it sometimes may not be (e.g., if the user is employing low-quality historical data). A failure mode arises when labels are fixed and highly uninformative, such that no SL iterate improves SSL initialization. In this case, vanilla SSL may be preferable. Second, we do not provide 
theoretical guarantees. The complexity of neural network training dynamics, particularly in settings with
multiple interacting components (e.g., projection/repair layers, PDEs) and in nonconvex optimization/simulation settings, would require restrictive assumptions and careful derivations. Nevertheless, our empirical analysis demonstrates the effectiveness of the framework across challenging problem settings. These limitations represent promising directions for future work.

\vspace{-0.5em}
\section{Related work}\label{sec:related}
\vspace{-0.5em}

\textbf{Amortized optimization.}
Amortized optimization trains ML models to map problem parameters directly to optimization solutions, replacing repeated iterative solves with fast forward inference \cite{amos2023tutorial}. Neural surrogates have achieved competitive optimality and feasibility while delivering orders-of-magnitude speedups in challenging nonconvex settings \cite{nguyen2025fsnet}. Many approaches rely on SL, regressing onto solutions produced by classical solvers, which yields stable training but incurs substantial offline labeling cost \cite{piloto2024canosfastscalableneural,lovett2024opfdatalargescaledatasetsac}. To reduce dependence on labeled data, recent work adopts self-supervised, objective-driven training. 
Constraint handling ranges from soft penalties \cite{fioretto2021lagrangian} to hard enforcement through projection or implicit layers with feasibility guarantees \cite{tordesillas2023rayen,  grontas2025pinet, liang2024homeomorphic, pan2022deepopf, donti2021dc3, nguyen2025fsnet, min2024hard, gould2016differentiating, agrawal2019differentiable, amos2017optnet}. While SSL approaches avoid expensive labels, they are often sensitive to initialization and conditioning in constrained, nonconvex problems. Our work bridges SL and SSL 
by using cheap-label supervision to warm-start SSL, improving optimality and constraint feasibility 
without requiring high-fidelity labels.

\textbf{Physics-informed machine learning.}
Physics-informed ML incorporates structural knowledge of governing dynamics to build surrogate solvers for differential equations. Physics-informed neural networks (PINNs) minimize PDE residuals at collocation points \cite{RAISSI2019686}, while neural operators such as the Fourier Neural Operator (FNO) learn resolution-invariant solution mappings \cite{li2021fourierneuraloperatorparametric}. However, the use of soft PDE-based regularization within loss functions leads to ill-conditioned optimization problems that exhibit challenging loss landscapes \cite{krishnapriyan2021characterizing}. Many physical systems admit multi-fidelity solutions, motivating approaches that learn corrections from low- to high-fidelity solvers \cite{um2021solverinthelooplearningdifferentiablephysics, li2023physicsinformedneuraloperatorlearning}. In contrast, we leverage approximations from linearized dynamics as structured pretraining data. These signals guide models toward physically plausible regions of the solution space before final SSL.

\textbf{Behavior cloning and reinforcement learning.}
Behavior cloning (BC) formulates policy learning as supervised learning from expert demonstrations but is brittle under noisy data and distribution shift. Common remedies fine-tune learned policies through online interaction with experts \cite{ross2011reduction}. Reinforcement learning (RL) instead uses rewards as a self-supervised signal to search for optimal policies, but typically requires careful reward shaping and large amounts of interaction data to be effective. Hybrid approaches combine BC and RL into a single offline objective by leveraging demonstration data and sparse rewards \cite{lu2023imitation}. Relatedly, several works successfully bootstrap RL from policies pretrained via BC \cite{silver2016mastering, rajeswaran2018learning} to reduce sample complexity. This two-stage pipeline is also reflective of how modern large models are trained, where generalized pretraining provides a strong initialization that is finetuned through objective-driven or preference-based optimization \cite{ouyang2022traininglanguagemodelsfollow, Lu2025-jl}. Inspired by these insights, our approach operates in the setting of amortized constrained optimization and studies how cheap labels can be used to warm-start and help SSL reach better minima faster.


\vspace{-0.5em}
\section{Conclusion}\label{sec:conclusion}
\vspace{-0.5em}

We present an intuitive yet principled framework for effective amortized optimization that bridges the strengths of supervised and self-supervised learning. Our key insight is that difficult nonconvex constrained problems do not require high-fidelity supervision, but rather initialization within a favorable regime for a final SSL stage. Based on this observation, we propose a three-stage pipeline that first collects a relatively small number of inaccurate labels, then trains a supervised model until achieving minimum merit loss, and then uses it to warm-start the final SSL. Through empirical analysis across synthetic nonconvex constrained optimization, power grid operation, and stiff physics-based simulation, we show that this cheap warm-starting strategy improves training stability, convergence speed, and final task performance, while substantially reducing offline computational cost. We further highlight the importance of the proposed merit loss and demonstrate that inexpensive labels therefore can effectively guide learning without overfitting to their biases. Overall, our framework is simple, modular, and compatible with existing amortized optimization methods, making it practical to adopt.

\newpage

\bibliography{references}
\bibliographystyle{unsrtnat}


\appendix

\renewcommand\thefigure{\thesection.\arabic{figure}}
\setcounter{figure}{0}

\renewcommand\thetable{\thesection.\arabic{table}}
\setcounter{table}{0}

\newpage


\section{Additional experimental results}\label{app:sec:results}

\begin{figure}[t]
    \centering
    \includestandalone[width=0.95\linewidth]{figures/make_basin_theorem1}
    \caption{
    \textbf{Weight-space intuition for cheap-label warm-starting.}
    Left: Fitting cheap labels can move the model parameters into a favorable region for subsequent self-supervised learning (SSL), such that the supervised endpoint $\theta_{\mathrm{SL}}$ can provide a useful warm start.
    Right: When continued supervised fitting overfits to cheap-label bias, the best SSL initialization may occur at an intermediate SL step $\theta_k$. We use the merit loss $\mathcal{M}$ to select this SL step before switching to SSL.
    }
    \label{fig:theorem_intuition}
\end{figure}

\input{figures/table_fsnet_from_penalty}

\subsection{Comparison with different variants}\label{app:ssec:variants}

Table~\ref{tab:socp:fsnet_from_penalty} compares several strategies for
FSNet under a fixed cheap-label budget (800 labels, 0.5s CPU time per instance).
We first consider two hybrid baselines that interleave supervised and
self-supervised signals within a single training phase.
\mt{Hybrid FSNet \textit{w/ cheap labels}} optimizes a composite loss at every
iteration, combining cheap-label regression with objective and penalty terms.
In contrast, \mt{Semi-Supervised FSNet \textit{w/ cheap labels}} separates these
signals across data instances: labeled samples are trained using supervised
losses, while unlabeled samples use self-supervised objective-and-penalty
losses.

We also evaluate two no-label warm-start variants that rely entirely on
self-supervision during pretraining, which can be viewed as curricula over SSL
objectives: \mt{FSNet \textit{w/ no-label warm-start} (feas.)} pretrains with feasibility terms only 
\begin{equation}
    \mathcal{L}_{\mathrm{SSL}}(\theta)
    = \mathbb{E}_{(x,\hat y)} \big[
        \rho\|c(\pi_\theta(x), x)\|^2  + R(\pi_\theta(x))
    \big],
\end{equation}
while \mt{FSNet \textit{w/ no-label warm-start} (both)} pretrains with both objective and feasibility terms \eqref{eq:ssl_obj}. Then we fine-tune each warm-start with \mt{FSNet}.

In contrast, \mt{FSNet \textit{w/ cheap-label warm-start} (Ours)} cleanly
separates the roles of data-driven supervised pretraining and self-supervised
optimization. This stage separation yields the best overall trade-off in
Table~\ref{tab:socp:fsnet_from_penalty}, achieving the lowest mean and worst-case
objective values while maintaining strong feasibility. These results reinforce
our central message: cheap labels are most effective when used for warm-starting,
after which SSL can reliably refine solutions. We conclude that: (1) Mixing supervised and self-supervised objectives within a
single training stage can introduce conflicting optimization signals,
especially when labels are inaccurate. (2) Pure SSL (including no-label
warm-starts or curricula) lacks access to the structured information encoded in
even imperfect datasets. (3) \mt{no-label warm-start (both)} encodes a similar optimization landscape as SSL, potentially leading to similar local minima.
Nevertheless, this observation suggests an interesting direction for future work: designing adaptive strategies that reconcile these paradigms and
enable effective single-stage training.

\input{figures/table_socp_solution_time}
\input{figures/table_socp_ssl_from_steps}


\begin{figure}[t!]
    \centering

    \begin{minipage}[t]{0.49\linewidth}
        \centering
        \includegraphics[width=0.9\linewidth, trim=0 -70 0 0, clip]{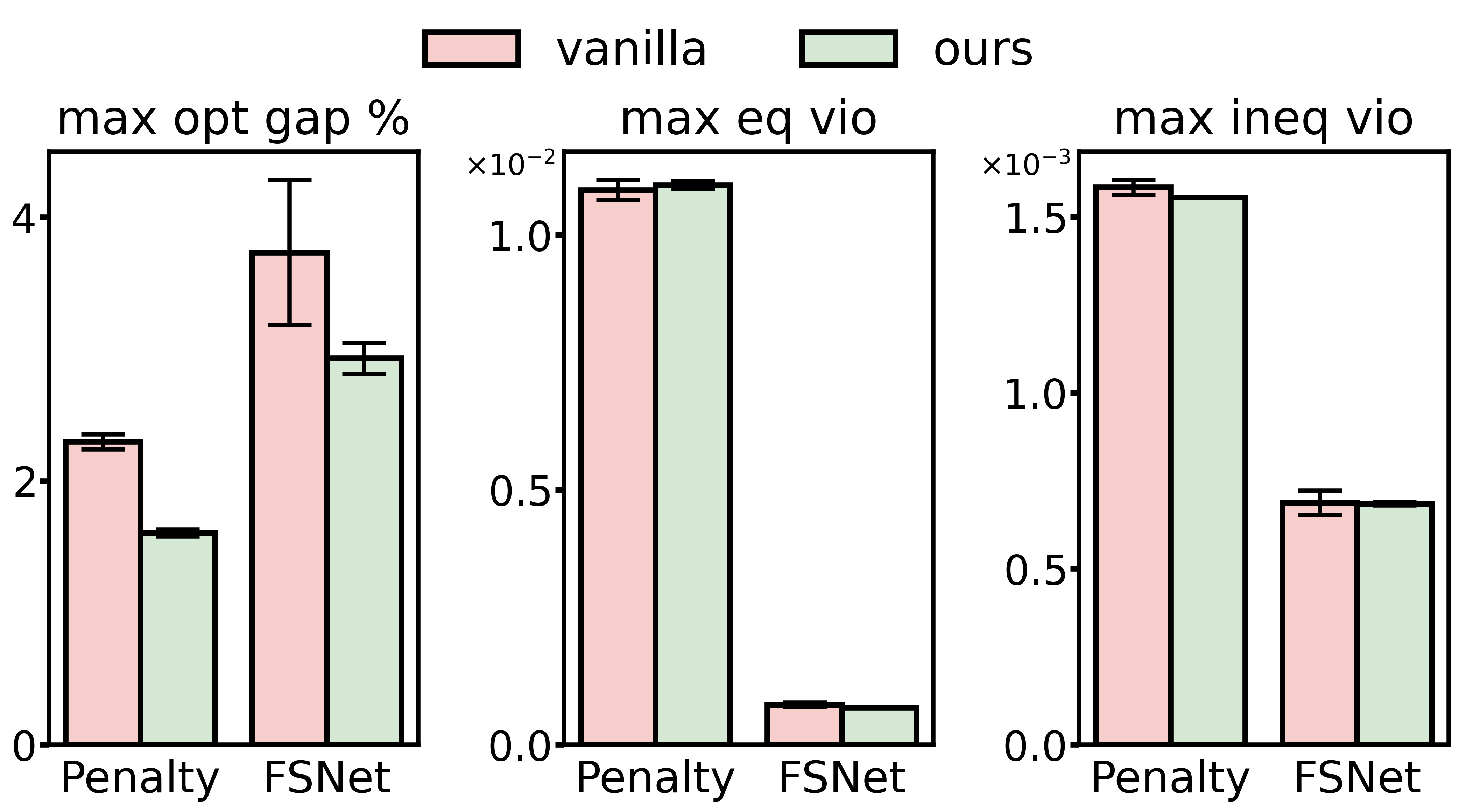}
        \caption{\textbf{Worst-case performance in learning optimal and feasible power grid operations.}}
        \label{fig:acopf:main_worst_case}
    \end{minipage}
    \hfill
    \begin{minipage}[t]{0.48\linewidth}
        \centering
        \includegraphics[width=\linewidth, trim=0 10 0 0, clip]{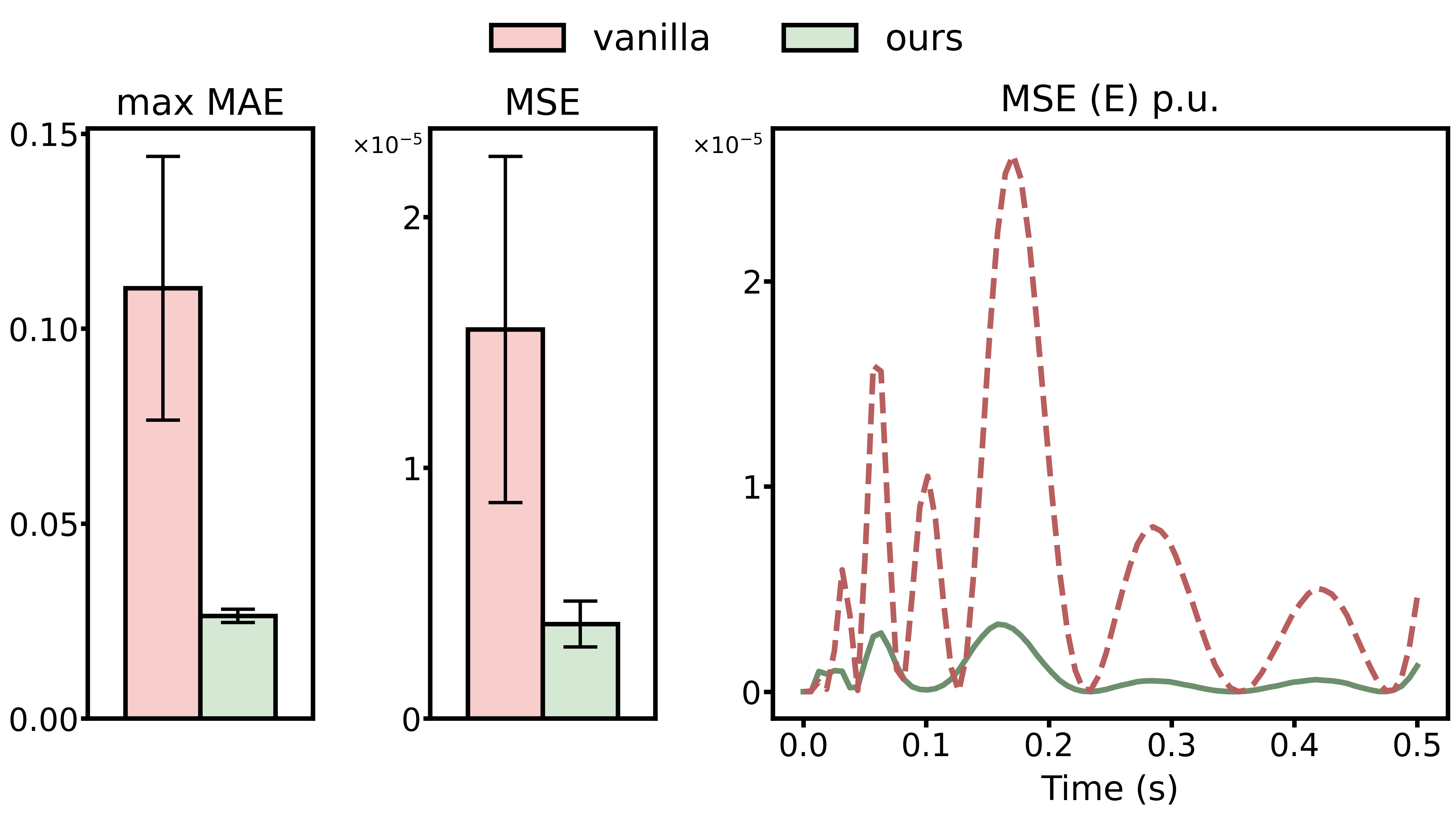}
        \caption{\textbf{Learning accurate physics simulation with R3 adaptive sampling method \cite{daw2022mitigating}.}}
        \label{fig:physics:basin_r3}
    \end{minipage}

\end{figure}

\begin{figure}[h!]
    \centering
    \includegraphics[width=0.8\linewidth]{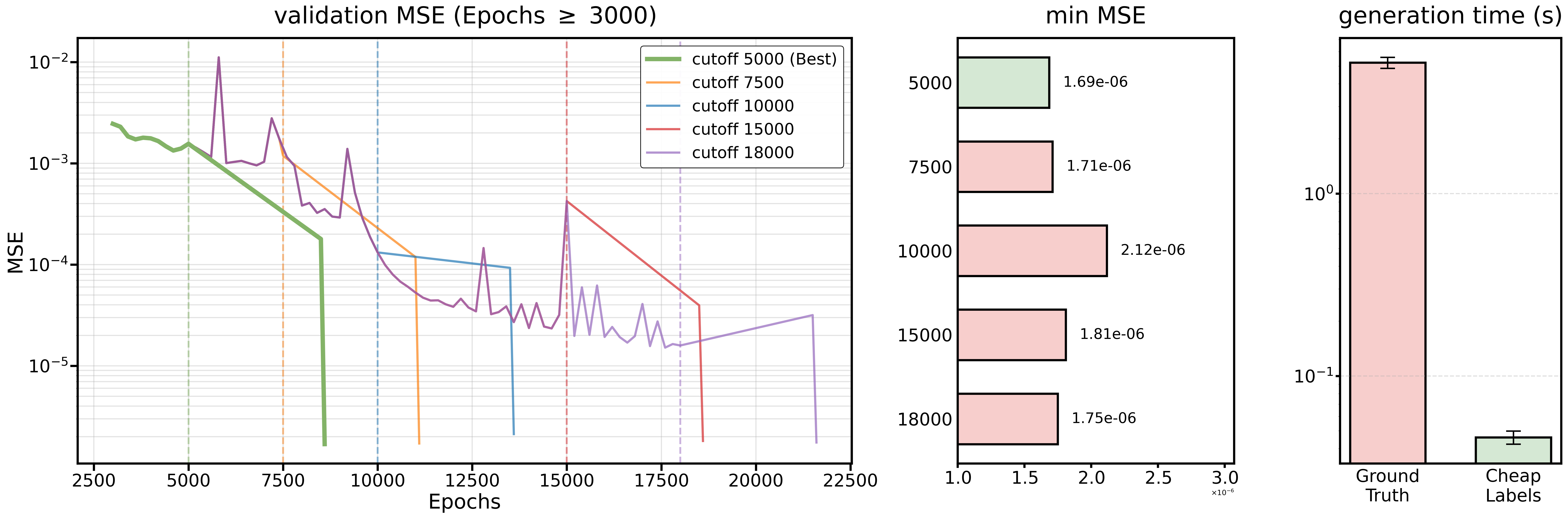}
    
    \caption{\textbf{Analysis of pre-training strategy and computational efficiency for RAR physics-informed learning.} 
    (a) Impact of warm-starting steps (left \& center): The learning curves (left) show the evolution of validation MSE, where the vertical lines mark the transition from warm-starting to self-supervised learning. The optimal cutoff at epoch 5,000 (green) leads to better convergence. The comparative bar chart (center) highlights that this optimal duration yields the lowest final error ($\approx 1.69 \times 10^{-6}$).
    (b) Label generation runtime (right): Our cheap labeling strategy achieves an $\sim$100$\times$ speedup compared to ground-truth generation.}
    \label{fig:physics:combined_analysis}
\end{figure}

\input{figures/table_socp_label_quality}
\input{figures/table_acopf_label_quality}

\section{Additional discussions}\label{app:sec:discussion}

\paragraph{The challenges of enforcing hard constraints.}

One promising approach for enforcing strict feasibility is to pass neural network outputs through iterative repair modules. However, training such systems end-to-end from scratch is difficult because the iterative repair process induces complicated training dynamics. Ideally, the ``plain neural network'' would first be trained in a self-supervised manner with a large penalty coefficient \(\rho\), producing predictions that are already close to feasible so that the repair module only needs to make small corrections. In practice, however, large \(\rho\) values lead to ill-conditioned or non-smooth optimization landscapes, making SSL difficult to train. As a result, SSL is typically performed with smaller \(\rho\) values to stabilize optimization, but this often yields predictions with larger constraint violations. Consequently, the repair module must apply larger corrections, further complicating end-to-end training dynamics. Our work addresses this challenge by using cheap labels and supervised pretraining to find a favorable initialization for subsequent end-to-end SSL training with iterative repair modules.

\paragraph{Relation to solver warm-starting.}

Our approach is related to but conceptually distinct from methods that learn solution initializations for iterative optimization solvers \cite{sambharya2024learning}.
In particular, our warm-starting scheme is focused on providing a useful initialization within the \emph{parameter space} of a neural network surrogate model (i.e., for the network weights~$\theta$) during model training; during inference, the neural network then predicts an optimization solution end-to-end, as standard for amortized optimization approaches.
In contrast, warm starting techniques for  optimization solvers focus on providing initializations within the \emph{solution space} of the problem; these solution-space initializations are then used to invoke a full iterative optimization technique in order to output a final solution to the optimization problem.
One interesting relationship between these approaches is that they both focus on using cheap, inexact solutions to improve the quality of the final solution; however, while our approach takes such inexact solutions as \emph{input} to use as labels during training, solver warm-starting approaches aim to \emph{output} inexact solutions to initialize an iterative solver.

\paragraph{Relation to neural network warm-starting.}
The benefits of warm-starting neural network training remain under debate. \citet{ash2020warm} observed that sequential warm-starting can degrade generalization due to imbalanced gradient contributions between old and new data, and proposed a ``shrink-and-perturb'' initialization scheme to mitigate this issue. In contrast, \citet{ahn2025revisiting} showed that the apparent generalization gap disappears under standard training practices, particularly with appropriate data augmentation. While these
works focus primarily on supervised continual learning, our setting is fundamentally different: we leverage a task-faithful merit loss function to \textit{regularize the warm start} and thereby \textit{align two distinct training phases}, supervised pretraining and self-supervised training, rather than repeatedly fine-tuning on evolving labeled datasets. This enables effective initialization and improved downstream generalization.

\paragraph{Relation to hybrid curricula.}
Our framework can alternatively be interpreted as a curriculum-based hybrid training strategy. In principle, Stages~2 and~3 can be merged into a single phase using a composite loss that combines supervised data-driven fitting terms with self-supervised objective-value and constraint-violation-penalty terms, governed by a weight curriculum. Training would initially emphasize supervised fitting, with the self-supervised objective and constraint terms inactive; once the task-level merit loss ceases to improve, the supervised fitting loss can be phased out while the self-supervised objective and constraint terms are progressively increased and any hard-constraint techniques are activated. 
However, our experiments (Appendix~\ref{app:ssec:variants}) show that naive hybrid curricula are less effective.


\subsection{Additional discussion of Remark \ref{rem:indicator}}\label{app:ssec:indicator}

Using the indicator function $I_{\mathcal{F}}$, the problem is equivalently
\[
\min_y \; f(y) + I_{\mathcal{F}}(y),
\]
where
\[
I_{\mathcal{F}}(y)=
\begin{cases}
0, & y\in\mathcal{F},\\
+\infty, & \text{otherwise}.
\end{cases}
\]

\noindent A number of different approaches are used to approximate the indicator function $I_{\mathcal{F}}$ in practice within optimization methods. 

\paragraph{Penalty methods.}
Quadratic penalty methods approximate $I_{\mathcal{F}}$ by
\[
\phi_\rho(y)=\rho\!\left(\|h(y)\|^2 + \|g(y)^+\|^2\right),
\]
and minimize $f(y)+\phi_\rho(y)$. Under standard constraint qualifications,
accumulation points of minimizers converge to KKT points as $\rho\to\infty$
\cite{nocedal2006numerical,bertsekas2014constrained}.

\paragraph{Barrier methods.}
Interior-point methods approximate $I_{\mathcal{F}}$ from within the feasible
region via
\[
\phi_\mu(y)=-\mu \sum_i \log(-g_i(y)),
\]
and minimize $f(y)+\phi_\mu(y)$ over strictly feasible points. Solutions converge to
KKT points as $\mu\to 0$ \cite{boyd2004convex}.

\paragraph{Augmented Lagrangian methods.}
Augmented Lagrangian formulations introduce dual variables $\lambda$ and minimize
\[
\mathcal{L}_\rho(y,\lambda)
=
f(y) + \lambda^\top c(y) + \tfrac{\rho}{2}\|c(y)\|^2,
\]
which improves conditioning relative to pure penalties while retaining exact
KKT consistency \cite{bertsekas2014constrained}.

\paragraph{Conditioning.}
Although these methods differ in numerical behavior, they all approximate the
indicator function $I_{\mathcal{F}}$. As the approximation becomes exact,
the objective develops sharp curvature or non-smoothness, explaining the
difficulty of directly optimizing such objectives with gradient-based learning.

\section{Experimental details}\label{app:sec:details}

\subsection{Parametric nonsmooth nonconvex program}\label{app:ssec:socp}

We evaluate our approach on a parametric nonsmooth, nonconvex optimization problem designed to stress both optimization stability and constraint satisfaction \citep{nguyen2025fsnet}. The problem takes the form
\begin{align}
     \min_{L \leq y \leq U}\;\; & \frac{1}{2} y^\top Q y + p^\top \sin(y) + \lambda \|y\|_2 \\
    \text{s.t.}\;\; & Ay = x, \qquad \| G_i \cos(y) + h_i \|_2 \leq c_i^\top y + d_i,
\end{align}
where $y \in \mathbb{R}^n$ denotes the decision variables and $x \in \mathbb{R}^m$ is a problem parameter.

The objective consists of a quadratic term, a nonlinear sinusoidal perturbation, and a nonsmooth $\ell_2$ regularization term. The constraints include linear equality constraints parameterized by $x$ and second-order cone constraints with nonlinear dependence on $y$ through elementwise cosine functions. Together, these components induce a highly nonconvex and nonsmooth optimization landscape with many local stationary points.

This problem is representative of optimization-first learning contexts in which classical solvers are sensitive to initialization and tolerance settings, and purely self-supervised training often exhibits instability. Unlike some existing methods that target specific problem families to leverage their particular structures \cite{grontas2025pinet, min2024hard}, our work aims to reveal and address the challenges associated with general formulations. In our experiments, we use this benchmark to study the effects of cheap-label warm starts under both soft and hard constraint enforcement, and to analyze convergence behavior, feasibility, and solution quality across different training regimes.

In Figure~\ref{fig:socp_loss_merit_landscape}, we train an \mt{FSNet} model using its
default training loss until convergence. From the converged weights, we
visualize both the training loss and the task-level merit loss landscapes along two
random directions in parameter space, inspired by \cite{li2018visualizing}. All evaluations are performed on a
held-out test set of 2{,}000 instances. 

In Table~\ref{tab:socp:main}, we pretrain a plain neural network with SL using 800 cheap
labels, each generated with a maximum CPU time of 0.5\,s. We then perform multiple SSL methods with
and without such warm-start. For SL with high-quality labels, we
use a distance-based training loss, whereas for SL with cheap labels, we
add an additional constraint-violation penalty.

In Figure~\ref{fig:socp:merit_loss_budgets} (left), we generate 7{,}000 labels at varying quality
levels (see Table~\ref{tab:socp:label_quality}), pretrain supervised warm-starts with
a violation penalty, and then perform FSNet training initialized from each
warm-start to study the effect of label quality on final task performance.

In Figure~\ref{fig:socp:merit_loss_budgets} (right), we vary the number of cheap labels
(up to 7{,}000, each generated with a maximum CPU time of 0.5\,s), pretrain
corresponding warm-starts, and evaluate FSNet training initialized from these
warm-starts to study the effect of cheap label quantity on final task performance.

In Table~\ref{tab:socp:fsnet_from_penalty}, we pretrain multiple warm-start models,
both with and without labeled data, and then perform FSNet training initialized
from each warm-start to study the impact of label availability and quality on
final performance.

In Figure~\ref{fig:socp:average_merit_epochs}, we pretrain a supervised warm-start using 7{,}000 labels generated with a maximum CPU time of 2.0\,s and a violation penalty. The SL merit loss reaches its minimum after approximately 250 epochs (evaluated on a validation set). From multiple SL steps (50, 100, 250, 500 and 600), we initiate FSNet training both from such warm-starts and from random initialization (vanilla) for 300 SSL epochs. This Figure demonstrates the training dynamics from SL to SSL. One SSL epoch is equal to a full pass over $D$ of problem input. One SL epoch is equal to a full pass over the supervised ``cheap labels'' dataset $\hat{D}$ of both input and labels. We count SL and SSL training epochs separately.

In Figure~\ref{fig:socp:merit_loss_correlation}, we pretrain a supervised warm-start using 7{,}000 labels generated with a maximum CPU time of 3.0\,s and a violation penalty. The SL merit loss reaches its minimum after approximately 250 epochs (evaluated on a validation set). From multiple SL points, we initiate FSNet SSL training both from such warm-starts and from random initialization (vanilla). We terminate FSNet SSL training early based on merit loss on a validation set and record the merit loss on a test set as final task performance for each SSL model (named according to its warm-starting SL step). This figure demonstrates better correlation between final task performance with SL merit loss than with SL training loss.

In Figure~\ref{fig:socp:merit_loss_landscape}, we compare the trained FSNet models with and without warm-starting. Inspired by \cite{li2018visualizing}, we visualize the merit loss landscape along a linear interpolation between the initial and final model parameters, $\theta(\alpha) = (1-\alpha)\theta_0 + \alpha\theta_1,$ highlighting differences in potential training trajectories induced by different strategies.

In Table~\ref{tab:offline_time}, we report the total offline time required by each
method to reach comparable performance to vanilla self-supervised learning,
highlighting the efficiency gains of our warm-start strategy.

\subsection{AC optimal power flow}\label{app:ssec:opf}

For the AC optimal power flow problem, we use the formulation from \cite{klamkin2025pglearn}, reproduced below:
\begin{align*}
    \min_{p^\mathrm{g}, q^\mathrm{g}, p^\mathrm{f}, q^\mathrm{f}, p^\mathrm{t}, q^\mathrm{t}, v, \theta} \quad & \sum_{i \in \mathcal{N}} \sum_{j \in \mathcal{G}_i} c_j p_j^\mathrm{g} \\
    \text{s.t.} \quad 
    & \sum_{j \in \mathcal{G}_i} p_j^\mathrm{g} - \sum_{j \in \mathcal{L}_i} p_j^\mathrm{d} - g_i^\mathrm{s} v_i^2 = \sum_{e \in \mathcal{E}_i} p^\mathrm{f}_e + \sum_{e \in \mathcal{E}_i^R} p_e^\mathrm{t} & \forall i \in \mathcal{N} \\
    & \sum_{j \in \mathcal{G}_i} q_j^\mathrm{g} - \sum_{j \in \mathcal{L}_i} q_j^\mathrm{d} + b_i^\mathrm{s} v_i^2 = \sum_{e \in \mathcal{E}_i} q^\mathrm{f}_e + \sum_{e \in \mathcal{E}_i^R} q_e^\mathrm{t} & \forall i \in \mathcal{N} \\
    & p_e^\mathrm{f} = g_e^\mathrm{tt} v_i^2 + g_e^\mathrm{ft} v_i v_j \cos(\theta_i - \theta_j) + b_e^\mathrm{ft} v_i v_j \sin(\theta_i - \theta_j) & \forall e = (i,j) \in \mathcal{E} \\
    & q_e^\mathrm{f} = -b_e^\mathrm{tt} v_i^2 - b_e^\mathrm{ft} v_i v_j \cos(\theta_i - \theta_j) + g_e^\mathrm{ft} v_i v_j \sin(\theta_i - \theta_j) & \forall e = (i,j) \in \mathcal{E} \\
    & p_e^\mathrm{t} = g_e^\mathrm{tt} v_j^2 + g_e^\mathrm{tf} v_i v_j \cos(\theta_i - \theta_j) - b_e^\mathrm{tf} v_i v_j \sin(\theta_i - \theta_j) & \forall e = (i,j) \in \mathcal{E} \\
    & q_e^\mathrm{t} = -b_e^\mathrm{tt} v_j^2 - b_e^\mathrm{tf} v_i v_j \cos(\theta_i - \theta_j) - g_e^\mathrm{tf} v_i v_j \sin(\theta_i - \theta_j) & \forall e = (i,j) \in \mathcal{E} \\
    & (p_e^\mathrm{f})^2 + (q_e^\mathrm{f})^2 \leq \overline{S}_e^2 & \forall e \in \mathcal{E} \\
    & (p_e^\mathrm{t})^2 + (q_e^\mathrm{t})^2 \leq \overline{S}_e^2 & \forall e \in \mathcal{E} \\
    & \underline{\Delta \theta}_e \leq \theta_i - \theta_j \leq \overline{\Delta \theta}_e & \forall e = (i,j) \in \mathcal{E} \\
    & \theta^{\text{ref}} = 0 \\
    & \underline{p}_i^\mathrm{g} \leq p_i^\mathrm{g} \leq \overline{p}_i^\mathrm{g} & \forall i \in \mathcal{G} \\
    & \underline{q}_i^\mathrm{g} \leq q_i^\mathrm{g} \leq \overline{q}_i^\mathrm{g} & \forall i \in \mathcal{G} \\
    & \underline{v}_i \leq v_i \leq \overline{v}_i & \forall i \in \mathcal{N} \\
    & -\overline{S}_e \leq p_e^\mathrm{f} \leq \overline{S}_e & \forall e \in \mathcal{E} \\
    & -\overline{S}_e \leq q_e^\mathrm{f} \leq \overline{S}_e & \forall e \in \mathcal{E} \\
    & -\overline{S}_e \leq p_e^\mathrm{t} \leq \overline{S}_e & \forall e \in \mathcal{E} \\
    & -\overline{S}_e \leq q_e^\mathrm{t} \leq \overline{S}_e & \forall e \in \mathcal{E}
\end{align*}
where $\cN$, $\cE$, and $\cG$ denote the sets of buses, branches, and generators, respectively; $p^\mathrm{g}$ and $q^\mathrm{g}$ represent the active and reactive generation powers; $p^\mathrm{f}$ and $ q^\mathrm{f}$ ($p^\mathrm{t}$ and $ q^\mathrm{t}$) denote the active and reactive power flows in the forward (reverse) direction; and $v$ and $\theta$ denote the voltage magnitude and angle. The network parameters $g^\mathrm{ff}, g^\mathrm{ft}, g^\mathrm{tf}, g^\mathrm{tt}$ and $b^\mathrm{ff}, b^\mathrm{ft}, b^\mathrm{tf}, b^\mathrm{tt}$ are obtained from the admittance matrix of the system.
In addition, $\overline{S}$ denotes the apparent power flow limit, while $\underline{\Delta \theta}$ and $\overline{\Delta \theta}$ specify the lower and upper bounds on the voltage angle difference. The parameters $\underline{p}^\mathrm{g}$ and $\overline{p}^\mathrm{g}$ ($\underline{q}^\mathrm{g}$ and $\overline{q}^\mathrm{g}$) indicate the lower and upper limits of active (reactive) generation, and $\underline{v}$ and $\overline{v}$ define the permissible range of voltage magnitudes.

\subsubsection{DC optimal power flow}\label{app:sssec:dcopf}

The DC optimal power flow (DCOPF) problem is a widely used linear approximation of ACOPF and underlies most electricity market operations. The DC approximation is motivated by several assumptions that are most accurate for transmission networks: (i) voltage magnitudes are fixed to one per-unit, (ii) voltage angle differences are small so that $\sin(\theta_i - \theta_j) \approx \theta_i - \theta_j$, and (iii) reactive power and line losses are neglected (zeros). 

Constraint~\eqref{eq:dc_balance} enforces nodal active power balance via Kirchhoff’s current law. Constraint~\eqref{eq:dc_flow} models active power flows using Ohm’s law. Constraint~\eqref{eq:dc_angle} bounds voltage angle differences across each branch, and \eqref{eq:dc_ref} fixes the reference bus angle. Finally, constraints~\eqref{eq:dc_gen} and \eqref{eq:dc_flow_limits} enforce generator and line flow limits.
\begin{align}
\min_{p^\mathrm{g},\, p^\mathrm{f},\, \theta} \quad
& \sum_{i \in \mathcal{N}} \sum_{j \in \mathcal{G}_i} c_j \, p_j^\mathrm{g} \label{eq:dc_obj} \\[4pt]
\text{s.t.} \quad
& \sum_{j \in \mathcal{G}_i} p_j^\mathrm{g}
- \sum_{j \in \mathcal{L}_i} p_j^\mathrm{d}
= \sum_{e \in \mathcal{E}_i} p_e^\mathrm{f}
+ \sum_{e \in \mathcal{E}_i^R} p_e^\mathrm{f},
&& \forall i \in \mathcal{N} \label{eq:dc_balance} \\[4pt]
& -b_e (\theta_i - \theta_j) - p_e^\mathrm{f} = 0,
&& \forall e = (i,j) \in \mathcal{E} \label{eq:dc_flow} \\[4pt]
& \underline{\Delta \theta}_e \le \theta_i - \theta_j \le \overline{\Delta \theta}_e,
&& \forall e = (i,j) \in \mathcal{E} \label{eq:dc_angle} \\[4pt]
& \theta^{\mathrm{ref}} = 0 \label{eq:dc_ref} \\[4pt]
& \underline{p}_i^\mathrm{g} \le p_i^\mathrm{g} \le \overline{p}_i^\mathrm{g},
&& \forall i \in \mathcal{G} \label{eq:dc_gen} \\[4pt]
& -\overline{S}_e \le p_e^\mathrm{f} \le \overline{S}_e,
&& \forall e \in \mathcal{E} \label{eq:dc_flow_limits}
\end{align}

Here, $\mathcal{N}$, $\mathcal{E}$, and $\mathcal{G}$ denote the sets of buses, branches, and generators, respectively. The variable $p^\mathrm{g}$ denotes active power generation, $p^\mathrm{f}$ denotes active power flow on each branch, and $\theta$ denotes bus voltage angles. The parameter $b_e$ is the branch susceptance, $\overline{S}_e$ is the line flow limit, and $\underline{\Delta \theta}_e$, $\overline{\Delta \theta}_e$ specify angle difference bounds.

To construct our DCOPF labels, we complete the missing target variables:  reactive power of zero and voltage magnitude of one. In Figure~\ref{fig:acopf:main}, we first pretrain a supervised warm-start model with a
violation penalty using 10{,}000 DCOPF labels. We then perform self-supervised
learning (SSL) for ACOPF both with and without this warm-start to assess its
impact on convergence and final performance.

\subsection{Physics-informed flow-map training}\label{app:ssec:physics}

\subsubsection{Model and flow-map parameterization}
We consider a four-state nonlinear dynamical system $x=[\delta,\omega,P_m,E]^\top$ governed by stiff swing--governor--exciter equations. Instead of learning a time-marching solver, we learn a flow map that predicts the state at any time given an initial condition. Let $T$ be the final horizon and define the normalized time $\tau=t/T\in[0,1]$. We shift and scale the state using an equilibrium $x_{\mathrm{eq}}$ and scale vector $s$ and define $y=(x-x_{\mathrm{eq}})/s$. The neural flow map is
\[
\widehat{\Phi}_\theta(\tau,y_0)=y_0+\tau\,\pi_\theta([\tau,y_0]),
\]
which enforces the initial condition exactly at $\tau=0$.

\subsubsection{Physics residual and loss}
Given $\hat{y}=\widehat{\Phi}_\theta(\tau,y_0)$, we recover $\hat{x}=x_{\mathrm{eq}}+s\odot\hat{y}$. We compute $\partial\hat{y}/\partial\tau$ using automatic differentiation and convert to physical time by
\[
\dot{\hat{y}}=\frac{1}{T_{\mathrm{eff}}}\frac{\partial\hat{y}}{\partial\tau},\qquad
\dot{\hat{x}}=s\odot\dot{\hat{y}},
\]
where $T_{\mathrm{eff}}$ is a curriculum horizon. The weighted residual is
\[
r(x)=
\begin{bmatrix}
\dot{\hat{\delta}}-\hat{\omega}\\
M\dot{\hat{\omega}}-(\hat{P}_m-P_e(\hat{\delta},\hat{E})-D\hat{\omega})\\
T_g\dot{\hat{P}}_m-(P_{\mathrm{ref}}-R\hat{\omega}-\hat{P}_m)\\
T_e\dot{\hat{E}}-(E_{\mathrm{ref}}-K_f\hat{\omega}-\hat{E})
\end{bmatrix},
\qquad r_y=r(x)\oslash s,
\]
where $\oslash$ denotes elementwise division. We define the normalized residual
$r_y \in \mathbb{R}^4$ componentwise as
\[
(r_y)_i = \frac{r_i(x)}{s_i}, \qquad i=1,\dots,4,
\]
which uses the same state scaling $s$ employed to define the normalized state
$y=(x-x_{\mathrm{eq}})/s$. This normalization removes unit imbalance across
equations and improves numerical conditioning of the physics loss.

We split residual energy into slow components $(\delta,\omega)$ and fast components $(P_m,E)$ and minimize
\[
\mathcal{L}_{\mathrm{phys}}=
w_{\mathrm{slow}}\mathbb{E}\|r_y^{(\delta,\omega)}\|_2^2+
w_{\mathrm{fast}}\mathbb{E}\|r_y^{(P_m,E)}\|_2^2,
\]
where $w_{\mathrm{slow}},w_{\mathrm{fast}}$ are set inversely proportional to current residual magnitudes and constrained by a floor on $w_{\mathrm{fast}}$ to preserve attention on stiff channels.

\subsubsection{Physics-informed training as amortized optimization}
The goal of training is to learn a single parametric operator that approximately satisfies the governing equations for all initial conditions, rather than solving a new time-integration problem for each trajectory.
We interpret physics-informed training as an instance of amortized constrained
optimization.
For a given initial condition $y_0$, the governing dynamics define an implicit
feasibility problem over trajectories $y(\tau)$:
\[
\text{find } y(\cdot)
\quad \text{s.t.} \quad
r_y(\tau;y)=0,\ \forall \tau\in[0,1], 
\qquad y(0)=y_0,
\]
where $r_y(\tau;y)$ denotes the normalized physics residual.
Solving this problem independently for each $y_0$ corresponds to repeatedly
running a numerical time integrator.

Amortized optimization replaces this family of per-instance solves with a single
parametric map
\[
y(\tau)\approx \widehat{\Phi}_\theta(\tau,y_0),
\]
shared across all initial conditions.
Training the neural operator therefore corresponds to learning a policy that
outputs approximately feasible solutions to the physics-constrained problem for
all $y_0$ drawn from a distribution.

We formalize this objective using an epigraph formulation.
Introducing a residual tolerance $\epsilon\ge 0$, we seek to solve:
\[
\begin{aligned}
\min_{\theta,\ \epsilon}\quad & \epsilon\\
\text{s.t.}\quad 
&\mathbb{E}_{\tau,y_0}\!\left[\|r_y(\tau;\theta,y_0)\|_2^2\right]\le \epsilon,\\
&\widehat{\Phi}_\theta(0,y_0)=y_0 ,
\end{aligned}
\]
which minimizes the smallest residual level $\epsilon$ such that the amortized
model satisfies the governing equations up to tolerance across the distribution
of initial conditions.

In practice, this constrained problem is solved via a quadratic-penalty
relaxation, yielding the physics-informed loss
\[
\mathcal{L}_{\mathrm{phys}}(\theta)
=
\mathbb{E}_{\tau,y_0}\!\left[\|r_y(\tau;\theta,y_0)\|_2^2\right],
\]
optionally augmented with adaptive or scale-aware weighting to improve
conditioning.
From this perspective, physics-informed learning does not aim to regress onto
precomputed trajectories, but rather to amortize the solution of a
physics-constrained optimization problem, replacing iterative solvers with a
single forward pass of a trained operator.

\subsubsection{Sampling and residual-based adaptive refinement}
Initial conditions are sampled uniformly from a hyper-rectangle around equilibrium in each state. At each iteration, we train on a mixture of uniformly drawn collocation points and an adaptive set constructed by residual-based adaptive refinement (RAR \cite{wu2023comprehensive}): periodically, we sample a large candidate set of $(\tau,y_0)$, compute $\|r_y\|_1$ for each, and add the top-$K$ hardest points to an auxiliary pool that is merged with the uniform set (with random truncation if the pool exceeds a fixed budget). We also evaluate our framework with Retain-Resample-Release sampling (R3) algorithm \cite{daw2022mitigating} (see Figure~\ref{fig:physics:basin_r3}).

\subsubsection{Training strategies}

\textbf{Stage 1: Warm-start pretraining.}
The warm-start phase combines the physics loss at the full horizon $T$ with two inexpensive structural guides:
(i) linearized flow targets around equilibrium, obtained by computing the Jacobian of the dynamics at the equilibrium $x_{\mathrm{eq}}$ and evaluating the exact linear solution
\[
x(t)=x_{\mathrm{eq}}+e^{At}(x_0-x_{\mathrm{eq}}), \qquad 
A=\left.\frac{\partial f}{\partial x}\right|_{x_{\mathrm{eq}}},
\]
and (ii) a tangent-matching loss that enforces the correct mass-weighted derivative at the initial time $\tau=0$.

The equilibrium $x_{\mathrm{eq}}$ is defined as a steady-state operating point satisfying $f(x_{\mathrm{eq}})=0$. It is obtained by solving a small set of algebraic equations arising from setting all time derivatives in the swing--governor--exciter model to zero. This computation is inexpensive and performed once per parameter setting. Linearizing the dynamics at $x_{\mathrm{eq}}$ yields a first-order Taylor approximation of the vector field, which is valid for small deviations from equilibrium. In power-systems terminology this regime is known as \emph{small-signal dynamics}, as it characterizes the response of the system to small perturbations around a nominal operating point. For readers outside power systems, this corresponds to the standard local linearization of a nonlinear dynamical system.

The linearized flow targets therefore encode correct local modes, damping, and stiffness structure near $x_{\mathrm{eq}}$, but are not intended to be globally accurate. The tangent-matching term complements this by enforcing that the learned flow map leaves each initial condition with the correct physical derivative, i.e., that $\dot{\hat{x}}(0)=f(x_0)$. Together, these guides shape the early optimization landscape and bias the parameters toward a favorable regime.

Both auxiliary losses are multiplied by time-varying weights that are smoothly annealed to zero over Stage~1. Consequently, by the end of Stage~1 the model is trained using only the physics loss, and Stage~2 continues optimization from this physics-consistent but well-conditioned initialization.

\textbf{Stage 2: Physics-only self-supervised learning.}
In Stage~2 we optimize only $\mathcal{L}_{\mathrm{phys}}$ using Adam together with a horizon curriculum
\[
T_{\mathrm{eff}}(k)=T\big(\alpha+(1-\alpha)(k/K)^p\big),
\]
which gradually increases the enforced time span. Uniform collocation samples are refreshed periodically, and RAR is interleaved to enrich the training set near high-residual regions.

After the Adam phase, we apply LBFGS at $T_{\mathrm{eff}}=T$. The optimizer is run in an outer loop until the relative decrease in $\mathcal{L}_{\mathrm{phys}}$ falls below a tolerance for several consecutive checks (plateau-based stopping). For comparisons, LBFGS can be evaluated on a fixed collocation set shared across runs.

\textbf{Vanilla PINN.}
The baseline removes Stage~1 entirely and trains from random initialization using exactly the same Stage~2 pipeline: same physics loss, same sampling and RAR/R3, same horizon curriculum, and same Adam and LBFGS schedules. The only difference between the proposed method and the baseline is the presence or absence of the warm-start phase.

\subsubsection{Evaluation protocol}
For each trained model, we generate test trajectories by evaluating the flow map at multiple times and initial conditions and compare them against high-accuracy RK4 solutions. Errors are reported as MSE, RMSE, and max-absolute error per state. To probe the nonconvex optimization landscape, we repeat training for multiple random seeds and declare a run \emph{degenerate} if its final error exceeds prescribed thresholds, thereby quantifying reliability of the two training strategies.

\subsubsection{Supervised learning with Huber loss}

Because labels are inaccurate, we implement supervised learning using the Huber loss, a robust regression loss that interpolates between squared
error and absolute error. For a residual $r$, it is defined as
\[
\ell_\delta(r) =
\begin{cases}
\tfrac{1}{2} r^2, & |r| \le \delta, \\
\delta\big(|r| - \tfrac{1}{2}\delta\big), & |r| > \delta,
\end{cases}
\]
where $\delta > 0$ controls the transition point. For small residuals, the Huber
loss behaves like an $\ell_2$ loss, encouraging smooth optimization, while for
large residuals it behaves like an $\ell_1$ loss, reducing sensitivity to
outliers.

\subsection{Hyperparameters}\label{app:ssec:hyperparams}

To facilitate reproducibility, we report the hyperparameter settings for the dataset, network architecture, optimizer, and each method in Tables~\ref{tab:hyperparams_1},~\ref{tab:hyperparams_2},~\ref{tab:hyperparams_flowmap_1}. It is important to note that $\lambda_{eq}$ and $\lambda_{ineq}$ correspond to $\rho$ of each type of constraints in SL and SSL training losses. \textbf{Anonymized code is included in the Supplementary Material and will be released publicly upon acceptance.}

\begin{table*}[t!]
    \centering
    \scriptsize
    \caption{Hyperparameter settings for synthetic optimization experiments.}
    \label{tab:hyperparams_1}
    
    \begin{subtable}{\textwidth}
        \centering
        \caption{Common Settings}
        \begin{tabular}{ll}
            \toprule
            \textbf{Parameter} & \textbf{Value} \\
            \midrule
            \multicolumn{2}{l}{\textit{Dataset}} \\
            Sample & 10000 \\
            Label & 10000 \\
            Test & 2000 \\
            Batch Size & 512 \\
            \midrule
            \multicolumn{2}{l}{\textit{Architecture}} \\
            Network & MLP (4 layers) \\
            Hidden Dim & 1024 \\
            Dropout & 0.1 \\
            \midrule
            \multicolumn{2}{l}{\textit{Optimizer}} \\
            Optimizer & AdamW \\
            Weight Decay & 0.001 \\
            Scheduler & Linear Warmup + Cosine Annealing \\
            Random Seeds & 0 1 2 3 \\
            \bottomrule
        \end{tabular}
    \end{subtable}

    \vspace{10pt} 

    \begin{subtable}{\textwidth}
        \centering
        \caption{Method-Specific Parameters}
        \begin{tabular}{lcccccc}
            \toprule
            \textbf{Method} & \textbf{LR} & \textbf{Ep.} & $\boldsymbol{\lambda_{obj}}$ & $\boldsymbol{\lambda_{eq}}$ & $\boldsymbol{\lambda_{ineq}}$ & \textbf{Specifics} \\
            \midrule
            \multicolumn{7}{l}{\textit{Supervised Baselines}} \\
            \mt{Supervised Compl} & $5\text{e-}4$ & 1000 & -- & -- & -- & Loss: Huber (Partial vars) \\
            \mt{Supervised} & $1\text{e-}4$ & 1000 & 0.1 & 10 & 10 & $\lambda_{sup}=100$ \\
            \midrule
            \multicolumn{7}{l}{\textit{Soft-Constraint SSL Baselines}} \\
            \mt{Penalty} & $1\text{e-}4$ & 1000 & 1.0 & 10 & 10 & -- \\
            \mt{Adaptive Penalty} & $1\text{e-}4$ & 1000 & 1.0 & $10{\to}500$ & $10{\to}100$ & Rate=2.0 \\
            \midrule
            \multicolumn{7}{l}{\textit{Hard-Constraint SSL Baselines}} \\
            \mt{DC3} & $5\text{e-}5$ & 1000 & 1.0 & 1.0 & 10 & Steps=20, $\eta_{corr}=1\text{e-}6$ \\
            \mt{FSNet} & $1\text{e-}4$ & 300 & 1.0 & 10 & 10 & $\lambda_{dist}=5$, Mem=30, Scl=1000 \\
            \mt{Hybrid FSNet} & $1\text{e-}4$ & 300 & 1.0 & 10 & 10 & $\lambda_{dist}=5$, $\lambda_{sup}=2$, Scl=500 \\
            \mt{Semi-Supervised FSNet} & $1\text{e-}4$ & 300 & 1.0 & 10 & 10 & Mem=30, Scl=1000 \\
            \bottomrule
        \end{tabular} \\
        \vspace{2pt}
         \textit{Note: ``Scl" denotes the scaling factor for the solver. ``Mem" is the L-BFGS memory size.}
        
    \end{subtable}
\end{table*}

\begin{table*}[t]
    \centering
    \scriptsize
    \caption{Hyperparameter settings for AC-OPF experiments.}
    \label{tab:hyperparams_2}
    
    \begin{subtable}[t]{0.35\textwidth}
        \centering
        \caption{Common Settings}
        \begin{tabular}{ll}
            \toprule
            \textbf{Parameter} & \textbf{Value} \\
            \midrule
            \multicolumn{2}{l}{\textit{Dataset}} \\
            Sample & 7000 \\
            Label & 50--7000 \\
            Val / Test & 1000 / 2000 \\
            \midrule
            \multicolumn{2}{l}{\textit{Architecture}} \\
            Network & 5-Layer MLP \\
            Activation & SiLU \\
            Hidden Size & 256 \\
            Dropout & 0.01 \\
            Post-Process & Bound \& Slack Repair \\
            \midrule
            \multicolumn{2}{l}{\textit{Optimizer}} \\
            Optimizer & Adam \\
            Betas & (0.9, 0.95) \\
            Weight Decay & $1 \times 10^{-5}$ \\
            Scheduler & Cosine w/ Warmup \\
            Random Seeds & 0 1 2 \\
            \bottomrule
        \end{tabular}
    \end{subtable}
    \hfill
    \begin{subtable}[t]{0.63\textwidth}
        \centering
        \caption{Method-Specific Parameters}
        \begin{tabular}{lccccc}
            \toprule
            \textbf{Method} & \textbf{LR} & \textbf{Epochs} & $\boldsymbol{\lambda_{eq}}$ & $\boldsymbol{\lambda_{ineq}}$ & $\boldsymbol{\lambda_{obj}}$ \\
            \midrule
            \mt{Supervised} & $5\text{e-}3^\dagger$ & 1000 & 2000 & 1000 & 1000 \\
            \mt{Penalty} & $1\text{e-}3$ & 1000 & 2000 & 1000 & $10 \cdot s_{obj}$ \\
            \mt{FSNet} & $2\text{e-}4$ & 130 & 2000 & 1000 & $10 \cdot s_{obj}$ \\
            \bottomrule
        \end{tabular}
        \vspace{2pt}

            \textit{Note: FSNet also uses $\lambda_{dist}=0.01$. $^\dagger$LR scaled by $\sqrt{\text{batch}/512}$.}

    \end{subtable}
\end{table*}

\subsubsection{PINN hyperparameters}

\begin{table*}[t]
    \centering
    \scriptsize
    \caption{Hyperparameter settings for flow-map PINN and local-minima experiments.}
    \label{tab:hyperparams_flowmap_1}
    
    \begin{subtable}[t]{0.35\textwidth}
        \centering
        \caption{Common Settings}
        \begin{tabular}{ll}
            \toprule
            \textbf{Parameter} & \textbf{Value} \\
            \midrule
            \multicolumn{2}{l}{\textit{System Setup}} \\
            States & $(\delta,\omega,P_m,E)$ (4D) \\
            Horizon $T$ & 0.5 \\
            IC domain & $\delta \!\pm\! 0.25$ \\
                       & $\omega \!\in\! [-0.3, 0.3]$ \\
                       & $P_m \!\pm\! 0.25$, $E \!\pm\! 0.25$ \\
            \midrule
            \multicolumn{2}{l}{\textit{Architecture}} \\
            Network & Residual flow-map MLP \\
            Hidden Layers & 3 \\
            Hidden Width & 64 \\
            Activation & Tanh \\
            \midrule
            \multicolumn{2}{l}{\textit{Optimizer}} \\
            Optimizer & Adam / LBFGS \\
            Precision & float64 \\
            Random Seeds & 0 1 2 3 4 \\
            \bottomrule
        \end{tabular}
    \end{subtable}
    \hfill
    \begin{subtable}[t]{0.63\textwidth}
        \centering
        \caption{Method-Specific Parameters}
        \begin{tabular}{lccc}
            \toprule
            \textbf{Stage} & \textbf{LR} & \textbf{Steps} & \textbf{Notes} \\
            \midrule
            Stage 1 (Pretrain) & $2 \times 10^{-3}$ & 15{,}000 & Physics + anchors \\
            Stage 2 (Physics) & $5 \times 10^{-4}$ & 3{,}500 & Physics only \\
            LBFGS Polish & -- & $\le 40$ & Final refinement \\
            Stage 2 Only & $5 \times 10^{-4}$ & 3{,}500 & No pretraining \\
            \bottomrule
        \end{tabular}
    \end{subtable}
\end{table*}

\textbf{Stage 1: Warm-start pretraining.}
Stage~1 combines physics-based residual minimization with auxiliary guidance terms to improve optimization stability and exploration. Collocation points are sampled uniformly in the joint space of normalized time $\tau\in[0,1]$ and normalized initial conditions $y_0$, using $n=3500$ points per iteration. To provide coarse trajectory supervision, anchor losses are computed from batch RK4 integrations, using $64$ initial conditions and $12$ time points per trajectory. In addition, linearized flow-map targets around the system equilibrium are incorporated by evaluating the exact matrix exponential of the Jacobian, again using $64$ initial conditions and $12$ time points.

To enforce local consistency at the initial time, a tangent loss is applied at $\tau=0$ using $512$ randomly sampled initial conditions. The physics residual loss employs a Huber penalty during early training, with threshold $\delta_0=2\times10^{-2}$, and transitions to a pure squared loss after $60\%$ of the training steps. The weights of the anchor and linearized losses are annealed using a cosine schedule
\[
a(s)=\tfrac{1}{2}\bigl(1+\cos(\pi s)\bigr),
\]
where $s$ denotes normalized training progress.

\textbf{Stage 2: Physics-only training with time curriculum and RAR.}
Stage~2 refines the learned flow-map using physics residuals only. Uniform collocation points are sampled with $n=4500$ points per iteration and refreshed every $200$ optimization steps. Training proceeds under a time-curriculum strategy in which the effective integration horizon is gradually increased according to
\[
T_{\mathrm{eff}} = T\bigl(\alpha + (1-\alpha)s^p\bigr),
\]
with $\alpha=0.2$ and $p=2$, where $s$ again denotes normalized training progress.

To focus optimization on difficult regions of the state--time domain, residual-based adaptive refinement (RAR) is applied for $12$ rounds. In each round, $400$ new collocation points are selected from $20000$ candidates based on the magnitude of the physics residual, with the total RAR set capped at $12000$ points. Stiff and non-stiff residual components are balanced adaptively using inverse-loss weighting, with a lower bound imposed on the fast-mode weight ($w_{\text{fast}}\ge0.25$) to ensure accurate learning of stiff dynamics.

\textbf{LBFGS polishing.}
Following Stage~2, the network parameters are further refined using the L-BFGS optimizer with strong Wolfe line search. The LBFGS objective corresponds to the balanced physics residual evaluated over the full time horizon $T$. Optimization is performed using an outer-loop convergence criterion with a relative improvement tolerance of $10^{-4}$ and a patience of $5$ iterations, where each outer iteration executes up to $50$ internal LBFGS steps. For comparison experiments, an optional fixed collocation set of $12000$ points is used to ensure identical LBFGS objectives across different training runs.

\textbf{PINN baseline.}
To isolate the effect of Stage~1 pretraining, a PINN baseline is trained using the same physics-only procedure, time curriculum, RAR and R3 strategies, and LBFGS polishing as above, but without any auxiliary losses or pretraining.


\newpage

\end{document}

%% file: preamble.tex
\def\cE{\mathcal{E}}  \def\cG{\mathcal{G}} 
   
 \def\cN{\mathcal{N}}

\usepackage{xspace}
\usepackage{helvet}
\usepackage{relsize}

\newcommand{\mt}[1]{{%
  \fontfamily{phv}\selectfont\scalebox{0.9}{#1}%
}\xspace}

\usepackage{lipsum}

\newcommand{\res}[2]{#1{\scriptsize$\pm$}#2}

\usepackage{array}
\usepackage{makecell}
\usepackage{xcolor,colortbl}

\definecolor{wscolor}{RGB}{240, 250, 240}

\newcolumntype{M}[1]{>{\raggedright\arraybackslash\setlength{\parskip}{0pt}\setlength{\parsep}{0pt}}p{#1}}

\theoremstyle{plain}
\newtheorem{theorem}{Theorem}[section]

\theoremstyle{definition}

\theoremstyle{remark}
\newtheorem{remark}[theorem]{Remark}

%% file: figures/table_exp_socp_main.tex
\begin{table*}[t!] 
\caption{\textbf{Performance comparison on a synthetic constrained optimization benchmark, } evaluated on 2{,}000 hold-out test instances, over four random seeds. Our variants that use cheap-label supervision 
consistently achieve lower objective values and improved feasibility, with reduced output variance compared to other learning-based baselines. Compared to the solver, amortized models solve 2{,}000 problem instances over $40{,}000\times$ faster with batched GPU inference, and over $100\times$ faster under sequential CPU execution (see Table~\ref{tab:socp_soln_time}).}
\label{tab:socp:main}

\centering
\begin{scriptsize}
\setlength{\tabcolsep}{6pt}
\renewcommand\cellalign{lc}

\scalebox{0.9}{
\begin{tabular}{M{3.1cm}ccccc}
\toprule

\multirow{2}{*}{\textbf{Method}}
& \multirow{2}{*}{\textbf{\makecell[c]{Mean\\ Objective $\downarrow$}}}
& \multicolumn{2}{c}{\textbf{Equality Violation} $\downarrow$}
& \multicolumn{2}{c}{\textbf{Inequality Violation} $\downarrow$} \\
\cmidrule(lr){3-4}\cmidrule(lr){5-6}
& & \textbf{Mean} & \textbf{Max} & \textbf{Mean} & \textbf{Max} \\
\midrule

\makecell[l]{\mt{Solver}}
& \res{-2.40}{0.00}
& \res{0.00}{0.00} & \res{0.00}{0.00}
& \res{3.40e-09}{0.00} & \res{5.90e-08}{0.00} \\

\midrule

\makecell[l]{\mt{Supervised w/ high-quality labels}}
& \res{-6.16}{0.06}
& \res{1.07e+01}{3.92e-02} & \res{1.85e+01}{6.12e-01}
& \res{7.16e+01}{8.22e-01} & \res{1.99e+02}{4.76e+00} \\

\addlinespace[2pt]

\makecell[l]{\mt{Supervised w/ cheap labels}}
& \res{22.49}{1.24}
& \res{1.90e+00}{4.44e-02} & \res{2.72e+00}{8.38e-02}
& \res{1.23e-01}{9.51e-02} & \res{2.41e+00}{1.54e+00} \\

\midrule

\makecell[l]{\mt{Penalty}}
& \res{-0.06}{1.26}
& \res{9.63e-01}{5.65e-02} & \res{1.54e+00}{2.06e-01}
& \res{2.72e-02}{9.68e-03} & \res{6.11e-01}{1.02e-01} \\

\addlinespace[1pt]

\rowcolor{wscolor}
\makecell[l]{\mt{Our Penalty}}
& \res{-3.29}{0.06}
& \res{9.56e-01}{3.69e-02} & \res{1.52e+00}{5.25e-02}
& \res{2.28e-02}{6.06e-03} & \res{5.89e-01}{6.59e-02} \\

\midrule

\makecell[l]{\mt{Adaptive Penalty}}
& \res{18.23}{2.90}
& \res{7.43e-01}{1.14e-02} & \res{1.06e+00}{1.73e-02}
& \res{4.05e-04}{8.98e-05} & \res{8.25e-02}{1.84e-02} \\

\addlinespace[2pt]

\rowcolor{wscolor}
\makecell[l]{\mt{Our Adaptive Penalty}}
& \res{9.79}{0.19}
& \res{8.39e-01}{2.83e-02} & \res{1.18e+00}{4.26e-02}
& \res{4.06e-04}{9.92e-05} & \res{9.28e-02}{1.69e-02} \\

\midrule

\makecell[l]{\mt{DC3}}
& \res{5.19}{3.07}
& \res{0.00}{0.00} & \res{0.00}{0.00}
& \res{1.71e-03}{1.07e-03} & \res{2.21e-01}{1.23e-01} \\

\addlinespace[2pt]

\rowcolor{wscolor}
\makecell[l]{\mt{Our DC3}}
& \res{-1.58}{0.05}
& \res{0.00}{0.00} & \res{0.00}{0.00}
& \res{9.24e-04}{2.69e-04} & \res{1.17e-01}{3.25e-02} \\

\midrule

\makecell[l]{\mt{FSNet}}
& \res{-0.73}{1.42}
& \res{5.85e-05}{2.42e-05} & \res{1.93e-03}{1.16e-03}
& \res{7.89e-07}{5.03e-07} & \res{1.31e-04}{1.62e-04} \\

\addlinespace[2pt]

\rowcolor{wscolor}
\makecell[l]{\mt{Our FSNet}}
& \res{-3.28}{0.14}
& \res{4.06e-05}{8.54e-06} & \res{1.67e-03}{8.96e-05}
& \res{5.01e-07}{5.91e-08} & \res{6.25e-05}{1.50e-05} \\

\bottomrule
\end{tabular}}
\end{scriptsize}
\vspace{-1em}
\end{table*}

%% file: figures/table_exp_offline_time.tex
\begin{table*}[t!]
\centering
\caption{\textbf{Breakdown of offline computational time  across label generation, supervised learning, and self-supervised learning.}
Pure SL is dominated by expensive label generation, while penalty-based SSL avoids labels but requires longer training. Our method uses a small number of cheap labels, substantially reducing labeling cost and accelerating SSL convergence, achieving comparable or better solution quality with orders-of-magnitude lower total offline time. For these timing comparisons, both vanilla and our methods are trained to the \textit{same} level of task performance.}
\label{tab:offline_time}
\begin{scriptsize}
\setlength{\tabcolsep}{6pt}
\begin{tabular}{M{5.5cm}cccc}
\toprule
 & \textbf{Gen (s)} & \textbf{SL (s)} & \textbf{SSL (s)} & \textbf{Total (s)} \\
\midrule
\multicolumn{5}{l}{\textbf{Synthetic optimization}} \\
\midrule
Supervised w/ 7{,}000 10.0 labels
& 28118 & 190 & 0 & 28308 \\
\addlinespace[4pt]
Penalty
& 0 & 0 & 180 & 180 \\
\rowcolor{wscolor} 
Penalty w/ warm-start via 800 0.5 labels (Ours)
& 400 & 40 & 40 & 480 \\
\addlinespace[4pt]
FSNet
& 0 & 0 & 990 & 990 \\
\rowcolor{wscolor}
FSNet w/ warm-start via 800 0.5 labels (Ours)
& 400 & 40 & 495 & 935 \\
\midrule
\multicolumn{5}{l}{\textbf{Power grid operations}} \\
\midrule
Supervised w/ 10{,}000 ACOPF labels 
& 827 & 400 & 0 & 1227\\
\addlinespace[4pt]
Penalty
& 0 & 0 & 370 & 370 \\
\rowcolor{wscolor}
Penalty w/ warm-start via 10{,}000 DCOPF labels (Ours)
& 34 & 400 & 300 & 734 \\
\addlinespace[4pt]
FSNet
& 0 & 0 & 16900 & 16900 \\
\rowcolor{wscolor}
FSNet w/ warm-start via 10{,}000 DCOPF labels (Ours)
& 34 & 400 & 9100 & 9534 \\
\bottomrule
\end{tabular}
\end{scriptsize}
\end{table*}

%% file: figures/table_socp_random_labels.tex
\begin{table*}[h!]
\centering
\caption{\textbf{Assessment of random-label warm-start on synthetic constrained optimization.}
We compare SSL performance with no warm-start, random-label warm-start, and cheap-label warm-start. Random solution labels are sampled uniformly over the cheap label ranges in each dimension.}
\label{tab:socp:ssl_random}

\scriptsize
\setlength{\tabcolsep}{4pt}
\renewcommand{\arraystretch}{1.02}
\renewcommand\cellalign{lc}

\begin{tabular}{M{3.0cm}ccc}
\toprule
\textbf{Method}
& \textbf{\makecell[c]{Mean Obj. $\downarrow$}}
& \textbf{\makecell[c]{Mean Eq. Vio. $\downarrow$}}
& \textbf{\makecell[c]{Mean Ineq. Vio. $\downarrow$}} \\
\midrule

Vanilla Penalty
& \res{-0.06}{1.26}
& \res{9.63e-01}{5.65e-02}
& \res{2.72e-02}{9.68e-03} \\

Penalty w/ random labels
& \res{3.89}{0.03}
& \res{1.03e+00}{3.81e-02}
& \res{3.26e-02}{9.31e-03} \\

\rowcolor{wscolor}
Penalty w/ cheap labels (Ours)
& \res{-3.29}{0.06}
& \res{9.56e-01}{3.69e-02}
& \res{2.28e-02}{6.06e-03} \\

\midrule

Vanilla FSNet
& \res{-0.73}{1.42}
& \res{5.85e-05}{2.42e-05}
& \res{7.89e-07}{5.03e-07} \\

FSNet w/ random labels
& \res{1.46}{0.68}
& \res{5.15e-05}{2.39e-05}
& \res{3.78e-06}{5.51e-06} \\

\rowcolor{wscolor}
FSNet w/ cheap labels (Ours)
& \res{-3.28}{0.14}
& \res{4.06e-05}{8.54e-06}
& \res{5.01e-07}{5.91e-08} \\

\bottomrule
\end{tabular}
\end{table*}

%% file: figures/make_basin_theorem1.tex
\begin{tikzpicture}[
    font=\sffamily,
    >=Stealth,
    card/.style={
        rectangle,
        draw=black!10,
        fill=white,
        rounded corners=8pt,
        drop shadow={opacity=0.10, shadow xshift=0.8ex, shadow yshift=-0.8ex},
        inner sep=13pt,
        align=center
    },
    good_region/.style={
        fill=cGood,
        fill opacity=0.10,
        draw=cGood,
        thick,
        dashed
    },
    sl_contour/.style={
        draw=cCheap!35,
        thin
    },
    traj/.style={
        line width=1.5pt,
        line cap=round
    },
    midarrow/.style={
        postaction={decorate, decoration={
            markings, mark=at position 0.55 with {\arrow[scale=1]{>}}
        }}
    },
    dot_ssl/.style={
        circle,
        fill=cGood,
        inner sep=2.4pt,
        label={[text=cGood!80!black, font=\bfseries\scriptsize, label distance=-10pt]above:$\theta^\star_{\mathrm{SSL}}$}
    },
    dot_sl/.style={
        circle,
        fill=cCheap,
        inner sep=2.4pt,
        label={[text=cCheap, font=\bfseries\scriptsize, label distance=2pt]above:$\theta_{\mathrm{SL}}$}
    }
]

\node[card] (case1) at (0,0) {
    \textbf{\textcolor{cCheap}{Case I: SL endpoint gives a good warm start}} \\[3pt]
    {\small \textcolor{cGray}{Cheap-label fitting ends in a favorable region for SSL.}}
    \vspace{0.6cm} \\

    \begin{tikzpicture}[scale=1.18]
        \useasboundingbox (-2.55,-2.25) rectangle (3.55,2.05);

        \coordinate (ssl) at (0,0);
        \coordinate (sl) at (0.85,0.35);
        \coordinate (start) at (-2.05,-1.0);
        \def\radius{1.45}

        \foreach \r in {0.45,0.85,1.25,1.65} {
            \draw[sl_contour] (sl) circle (\r);
        }

        \fill[good_region] (ssl) circle (\radius);
        \draw[good_region] (ssl) circle (\radius);
        \node[text=cGood!80!black, font=\scriptsize, align=center] at (-1.15,1.5)
        {favorable\\SSL region};

        \draw[traj,cCheap,midarrow]
        (start) .. controls (-0.55,-2.05) and (1.45,-0.95) .. (sl);

        \node[circle, fill=cGray, inner sep=1.6pt,
              label={[text=cGray, font=\scriptsize, label distance=2pt]left:$\theta_0$}] at (start) {};

        \node[dot_ssl] at (ssl) {};
        \node[dot_sl] at (sl) {};

        \node[text=cGray, font=\tiny, anchor=south east] at (3.45,-2.12)
        {2D slice of parameter space};
    \end{tikzpicture}
};

\node[card, right=0.8cm of case1] (case2) {
    \textbf{\textcolor{cCheap}{Case II: Merit selects an earlier checkpoint}} \\[3pt]
    {\small \textcolor{cGray}{The best SSL warm start may occur before the SL endpoint.}}
    \vspace{0.6cm} \\

    \begin{tikzpicture}[scale=1.18]
        \useasboundingbox (-2.55,-2.25) rectangle (3.55,2.05);
        \clip (-2.45,-2.10) rectangle (3.45,1.92);

        \coordinate (ssl) at (0,0);
        \coordinate (sl) at (2.45,0.0);
        \coordinate (start) at (-1.85,-1.15);
        \coordinate (checkpoint) at (0.35,-0.70);
        \def\radius{1.18}

        \foreach \r in {0.5,1.0,1.5,2.0,2.5,3.0} {
            \draw[sl_contour] (sl) circle (\r);
        }

        \fill[good_region] (ssl) circle (\radius);
        \draw[good_region] (ssl) circle (\radius);
        \node[text=cGood!80!black, font=\scriptsize, align=center] at (-0.93,1.45)
        {favorable\\SSL region};

        \draw[traj,cGood,midarrow]
        (start) .. controls (-1.0,-2.05) and (-0.15,-1.15) .. (checkpoint);

        \draw[traj,cDanger,dashed,->]
        (checkpoint) .. controls (0.95,-0.20) and (1.75,0.03) .. (sl);

        \node[circle, fill=cGray, inner sep=1.6pt,
              label={[text=cGray, font=\scriptsize, label distance=-10pt]left:$\theta_0$}] at (start) {};

        \node[cross out, draw=cStop, line width=1.8pt, inner sep=3.1pt] (stop) at (checkpoint) {};
        \node[text=cStop, font=\scriptsize\bfseries, below=4pt of stop]
        {$\theta_k$};
        \node[text=cStop, font=\tiny, align=center, below=14pt of stop]
        {lowest merit loss};

        \node[dot_ssl] at (ssl) {};
        \node[dot_sl] at (sl) {};

        \node[font=\scriptsize, text=cDanger, anchor=west, align=left] at (1.20,1.02)
        {continued SL fitting\\follows label bias};

        \node[text=cGray, font=\tiny, anchor=south east] at (3.38,-2.02)
        {2D slice of parameter space};
    \end{tikzpicture}
};

\end{tikzpicture}

%% file: figures/table_fsnet_from_penalty.tex
\begin{table*}[h!]
\caption{\textbf{Performance comparison of FSNet with different training schemes on a synthetic constrained optimization benchmark (800 labels, max CPU time 0.5s).}
We report the mean and maximum objective value, $\ell_1$ equality violation, and $\ell_1$ inequality violation for vanilla FSNet and FSNet with different warm-start strategies. The pretraining loss in (feas.) only includes feasibility and (both) additionally includes objective. Lower is better.}
\label{tab:socp:fsnet_from_penalty}

\centering
\begin{scriptsize}
\setlength{\tabcolsep}{5pt}
\renewcommand\cellalign{lc}
\renewcommand{\arraystretch}{1.08}

\begin{tabular}{M{4.3cm}ccc}
\toprule

\multirow{2}{*}{\textbf{Method}}
& \multirow{2}{*}{\textbf{\makecell[c]{Mean \\ Objective $\downarrow$}}}
& \textbf{Equality Violation} $\downarrow$
& \textbf{Inequality Violation} $\downarrow$ \\

& & \textbf{Mean (Max)} & \textbf{Mean (Max)} \\

\midrule

\makecell[l]{\mt{FSNet}}
&
\res{-0.73}{1.42}
&
\makecell{
\res{5.85e{-05}}{2.42e{-05}} \\
(\res{1.93e{-03}}{1.16e{-03}})
}
&
\makecell{
\res{7.89e{-07}}{5.03e{-07}} \\
(\res{1.31e{-04}}{1.62e{-04}})
}
\\

\addlinespace[2pt]

\makecell[l]{\mt{Hybrid FSNet \textit{w/ cheap labels}}}
&
\res{2.68}{4.55}
&
\makecell{
\res{7.17e{-05}}{9.49e{-06}} \\
(\res{1.11e{-03}}{1.85e{-04}})
}
&
\makecell{
\res{4.12e{-07}}{1.00e{-07}} \\
(\res{3.35e{-05}}{1.42e{-05}})
}
\\

\addlinespace[2pt]

\makecell[l]{\mt{Semi-Supervised FSNet \textit{w/ cheap labels}}}
&
\res{-0.33}{0.50}
&
\makecell{
\res{8.84e{-05}}{2.36e{-05}} \\
(\res{1.00e{-03}}{2.94e{-04}})
}
&
\makecell{
\res{5.43e{-07}}{1.23e{-07}} \\
(\res{3.48e{-05}}{1.17e{-05}})
}
\\

\addlinespace[2pt]

\makecell[l]{\mt{FSNet \textit{w/ no-label warm-start} (feas.)}}
&
\res{-2.99}{0.01}
&
\makecell{
\res{9.97e{-05}}{2.87e{-06}} \\
(\res{8.39e{-04}}{7.14e{-05}})
}
&
\makecell{
\res{5.05e{-07}}{4.04e{-08}} \\
(\res{2.14e{-05}}{1.06e{-06}})
}
\\

\addlinespace[2pt]

\makecell[l]{\mt{FSNet \textit{w/ no-label warm-start} (both)}}
&
\res{0.30}{0.03}
&
\makecell{
\res{8.35e{-05}}{6.81e{-06}} \\
(\res{1.13e{-03}}{1.16e{-04}})
}
&
\makecell{
\res{4.86e{-07}}{6.31e{-08}} \\
(\res{3.62e{-05}}{4.31e{-06}})
}
\\

\addlinespace[2pt]

\rowcolor{wscolor}
\makecell[l]{\mt{FSNet \textit{w/ cheap-label warm-start} (Ours)}}
&
\res{-3.28}{0.14}
&
\makecell{
\res{4.06e{-05}}{8.54e{-06}} \\
(\res{1.67e{-03}}{8.96e{-05}})
}
&
\makecell{
\res{5.01e{-07}}{5.91e{-08}} \\
(\res{6.25e{-05}}{1.50e{-05}})
}
\\

\bottomrule
\end{tabular}
\end{scriptsize}
\end{table*}

%% file: figures/table_socp_solution_time.tex
\begin{table}[h!]
\caption{\textbf{Comparison of solution time at inference on the synthetic constrained optimization benchmark.}
Compared to the solver (IPOPT), amortized models solve 2{,}000 instances over $40{,}000\times$ faster with batched GPU inference, and over $100\times$ faster under sequential CPU execution. Our framework preserves such runtime benefits.}
\vspace{0.5em}
\label{tab:socp_soln_time}

\centering
\begin{scriptsize}
\setlength{\tabcolsep}{6pt}
\renewcommand\cellalign{lc}
\renewcommand{\arraystretch}{1.08}

\begin{tabular}{M{3.2cm}cc}
\toprule

\multirow{2}{*}{\textbf{Method}}
& \multicolumn{2}{c}{\textbf{Solution Time (s)} $\downarrow$} \\

\cmidrule(lr){2-3}

& \textbf{Batched GPU} & \textbf{Sequential CPU} \\

\midrule

\makecell[l]{\mt{Solver}}
& --
& \res{8033.60}{0.00} \\

\midrule

\makecell[l]{\mt{Supervised w/ high-quality data}}
& \res{0.0003}{0.0000}
& \res{0.9479}{0.0351} \\

\addlinespace[2pt]

\makecell[l]{\mt{Supervised w/ cheap data}}
& \res{0.0003}{0.0000}
& \res{0.9479}{0.0351} \\

\midrule

\makecell[l]{\mt{Penalty}}
& \res{0.0003}{0.0000}
& \res{0.9479}{0.0351} \\

\rowcolor{wscolor}
\makecell[l]{\mt{Our Penalty}}
& \res{0.0003}{0.0000}
& \res{0.9479}{0.0351} \\

\midrule

\makecell[l]{\mt{Adaptive Penalty}}
& \res{0.0003}{0.0000}
& \res{0.9479}{0.0351} \\

\rowcolor{wscolor}
\makecell[l]{\mt{Our Adaptive Penalty}}
& \res{0.0003}{0.0000}
& \res{0.9479}{0.0351} \\

\midrule

\makecell[l]{\mt{DC3}}
& \res{0.0960}{0.0000}
& \res{32.0881}{0.0280} \\

\rowcolor{wscolor}
\makecell[l]{\mt{Our DC3}}
& \res{0.0959}{0.0000}
& \res{31.9210}{0.0588} \\

\midrule

\makecell[l]{\mt{FSNet}}
& \res{0.1970}{0.0001}
& \res{80.0849}{0.2201} \\

\rowcolor{wscolor}
\makecell[l]{\mt{Our FSNet}}
& \res{0.2014}{0.0015}
& \res{71.9913}{0.4975} \\

\bottomrule
\end{tabular}

\end{scriptsize}
\end{table}

%% file: figures/table_socp_ssl_from_steps.tex
\begin{table*}[h!]
\caption{\textbf{Performance comparison on a synthetic constrained optimization benchmark of SSL warm-started from different SL pretraining steps.} We use $7,000$ inexact labels (with a maximum CPU solve time of 2.0 s per label) for supervised warm-starting. SL achieves lowest validation merit at around epoch 250. We then perform FSNet SSL for 300 epochs from different SL pretraining steps. Results are reported over 4 random seeds.}
\label{tab:socp:ssl_steps}
\centering
\begin{scriptsize}
\setlength{\tabcolsep}{6pt}
\renewcommand\cellalign{lc}
\begin{tabular}{M{2.33cm}ccccc}
\toprule
\multirow{2}{*}{\textbf{Method}}
& \multirow{2}{*}{\textbf{\makecell[c]{Mean\\ Objective $\downarrow$}}}
& \multicolumn{2}{c}{\textbf{Equality Violation} $\downarrow$}
& \multicolumn{2}{c}{\textbf{Inequality Violation} $\downarrow$} \\
\cmidrule(lr){3-4}\cmidrule(lr){5-6}
& & \textbf{Mean} & \textbf{Max} & \textbf{Mean} & \textbf{Max} \\
\midrule

\makecell[l]{{Vanilla SSL}}
& \res{-0.73}{1.42}
& \res{5.85e-05}{2.42e-05} & \res{1.93e-03}{1.16e-03}
& \res{7.89e-07}{5.03e-07} & \res{1.31e-04}{1.62e-04} \\
\addlinespace[3pt]

\makecell[l]{{SSL from SL step 50}}
& \res{-1.32}{0.04}
& \res{6.69e-05}{1.39e-05} & \res{1.35e-03}{1.43e-04}
& \res{4.60e-07}{1.47e-07} & \res{4.89e-05}{1.54e-05} \\
\addlinespace[3pt]

\makecell[l]{{SSL from SL step 100}}
& \res{-1.28}{1.12}
& \res{2.17e-05}{6.56e-06} & \res{2.32e-03}{6.21e-04}
& \res{4.79e-07}{2.70e-07} & \res{9.71e-05}{6.03e-05} \\
\addlinespace[3pt]

\rowcolor{wscolor}
\makecell[l]{{SSL from SL step 250*}}
& \res{-2.60}{0.16}
& \res{2.16e-05}{4.51e-06} & \res{1.89e-03}{1.77e-04}
& \res{3.32e-07}{3.16e-08} & \res{6.30e-05}{1.41e-05} \\
\addlinespace[3pt]


\makecell[l]{{SSL from SL step 500}}
& \res{-0.05}{0.19}
& \res{4.70e-05}{3.59e-05} & \res{7.85e-03}{8.98e-03}
& \res{6.35e-06}{9.02e-06} & \res{1.55e-03}{2.28e-03} \\
\addlinespace[3pt]

\makecell[l]{{SSL from SL step 600}}
& \res{0.05}{0.50}
& \res{3.88e-05}{1.54e-05} & \res{6.60e-03}{4.81e-03}
& \res{3.69e-06}{3.01e-06} & \res{9.04e-04}{7.79e-04} \\


\bottomrule
\end{tabular}
\end{scriptsize}
\end{table*}

%% file: figures/table_socp_label_quality.tex
\begin{table}[h!]
\centering
\caption{\textbf{Label generation cost and quality for synthetic constrained optimization.}
The maximum CPU time allowed per instance controls both the number of solver iterations and the total wall-clock time required to generate 10{,}000 labels, which we use to vary label quality. Allowing 10.0\,s per instance ensures solver convergence for all problems under tight tolerances, yielding the highest-quality labels collected.}
\vspace{0.5em}
\label{tab:socp:label_quality}
\begin{scriptsize}
\setlength{\tabcolsep}{6pt}
\begin{tabular}{cccccccc}
\toprule
\textbf{Max CPU Time}
& \textbf{Obj.}
& \textbf{Eq.\ Vio.}
& \textbf{Ineq.\ Vio.\ }
& \textbf{Success Rate}
& \textbf{Total Time (s)}
& \textbf{\# Iters}
& \textbf{Merit}\\
\midrule
0.5  &  1.06  & 3.77 & 122.23 & 0.00 &  4957  &  14  & $1.26\times10^{7}$ \\
1.5  & 11.54  & 1.31 &  39.53 & 0.00 & 14712  &  46  & $4.08\times10^{6}$ \\
2.0  & 19.33  & 0.37 &  11.68 & 0.00 & 20265  &  64  & $1.21\times10^{6}$ \\
3.0  & 10.32  & 0.01 &   1.05 & 0.03 & 30229  &  95  & $1.06\times10^{5}$ \\
4.0  & -1.67  & 0.00 &   0.21 & 0.56 & 37845  & 122  & $2.10\times10^{4}$ \\
10.0* & -2.44  & 0.00 &   0.00 & 1.00 & 40168  & 128  & $-2.44\times10^{0}$ \\
\bottomrule
\end{tabular}
\end{scriptsize}
\end{table}

%% file: figures/table_acopf_label_quality.tex
\begin{table}[h!]
\centering
\caption{\textbf{Comparison of ACOPF and DCOPF data generation.} We report normalized average objective (ACOPF being 1.00), $\ell_2$ constraint violation, merit value and total computation time of producing 10,000 labels. We solve both formulations to a tight tolerance without any budget constraints.}
\vspace{0.5em}
\label{tab:acopf:label_quality}
\begin{scriptsize}
\setlength{\tabcolsep}{6pt}
\begin{tabular}{cccccc}
\toprule
\textbf{Method}
& \textbf{Normalized Obj.}
& \textbf{Eq.\ Violation}
& \textbf{Ineq.\ Violation}
& \textbf{Total Time (s)}
& \textbf{Merit} \\
\midrule
ACOPF ($10^{-6}$)
& 1.000
& $5.0\times10^{-4}$
& $1.743\times10^{-6}$
& 827
& $5.12\times10^{1}$ \\
DCOPF
& 0.957
& $4.978\times10^{1}$
& $8.265\times10^{-7}$
& 34
& $4.98\times10^{6}$ \\
\bottomrule
\end{tabular}
\end{scriptsize}
\end{table}